\documentclass[10pt,journal,compsoc,twoside]{IEEEtran}
\usepackage{ifpdf}

  \usepackage{cite}
\ifCLASSINFOpdf
   \usepackage[pdftex]{graphicx}
   \graphicspath{{../pdf/}{../jpeg/}}
   \DeclareGraphicsExtensions{.pdf,.jpeg,.png}
\else
   \usepackage[dvips]{graphicx}
   \graphicspath{{../eps/}}
   \DeclareGraphicsExtensions{.eps}
\fi
\usepackage{amsmath,bm,amsfonts}
\usepackage{algorithmic}

\usepackage{array}

\ifCLASSOPTIONcompsoc
  \usepackage[caption=false,font=footnotesize,labelfont=sf,textfont=sf]{subfig}
\else
  \usepackage[caption=false,font=footnotesize]{subfig}
\fi

\usepackage{stfloats}
\usepackage{graphicx}
\usepackage{color}

\usepackage{mathrsfs}
\usepackage{makecell,multirow,diagbox}

\usepackage{algorithm}

\usepackage{makecell}

\usepackage{ragged2e}

\begin{document}
%
\title{Statistical Analysis of Signal-Dependent Noise: Application in Blind Localization of Image Splicing Forgery}
%
%
%
%

\author{Mian~Zou,~Heng~Yao,~\IEEEmembership{Member,~IEEE},~Chuan~Qin,~Xinpeng~Zhang,~\IEEEmembership{Member,~IEEE}
\IEEEcompsocitemizethanks{\IEEEcompsocthanksitem M. Zou is with School of Mechanical Engineering,
	University of Shanghai for Science and Technology, Shanghai 200093, China.\protect\\
	E-mail: chnzm366aq@163.com.
	\IEEEcompsocthanksitem H.~Yao and C.~Qin are with School of Optical-Electrical and Computer Engineering,
	University of Shanghai for Science and Technology, Shanghai 200093, China.\protect\\
	E-mail: hyao@usst.edu.cn; qin@usst.edu.cn.
    \IEEEcompsocthanksitem X. Zhang is with School of Computer Science, Fudan University, Shanghai 200433, China.\protect\\
    E-mail: zhangxinpeng@fudan.edu.cn.}
    \thanks{Manuscript received Oct 30, 2020. This work was supported by the National Natural Science Foundation of China (61702332, 61672354, U1936214, U1636206, and 61525203.) (\emph{Corresponding author: Heng~Yao}.)}}

%
%

\markboth{}%
{zou \MakeLowercase{\textit{et al.}}: Statistical Analysis of Signal-Dependent Noise: Application in Blind Localization of Image Splicing Forgery}
%



\IEEEtitleabstractindextext{
\begin{abstract}
\justifying
Visual noise is often regarded as a disturbance in image quality, whereas it can also provide a crucial clue for image-based forensic tasks. Conventionally, noise is assumed to comprise an additive Gaussian model to be estimated and then used to reveal anomalies. However, for real sensor noise, it should be modeled as signal-dependent noise (SDN). In this work, we apply SDN to splicing forgery localization tasks. Through statistical analysis of the SDN model, we assume that noise can be modeled as a Gaussian approximation for a certain brightness and propose a likelihood model for a noise level function. By building a maximum \emph{a posterior} Markov random field (MAP-MRF) framework, we exploit the likelihood of noise to reveal the alien region of spliced objects, with a probability combination refinement strategy. To ensure a completely blind detection, an iterative alternating method is adopted to estimate the MRF parameters. Experimental results demonstrate that our method is effective and provides a comparative localization performance.
\end{abstract}

}

\maketitle

\IEEEdisplaynontitleabstractindextext

%
\IEEEpeerreviewmaketitle

\section{Introduction}\label{introduction}

%
%
%
%
\IEEEPARstart{D}{uring} the process of acquisition or transmission, visual signals are often distorted by various types of disturbance, including noise, blur, enhancement, etc. Among these, noise is among the most common. Moreover, noise is inevitably incurred during visual information acquisition \cite{bovik}, such as digital photography under low light condition. Many works have explored how to estimate the noise and remove noise to improve visual quality. However, noise is not always useless; for example, it is beneficial in image forensics, especially in image splicing forgery localization, when it does not interfere with the visual experience. As is generally known, the wide availability of powerful image editing tools has made image manipulation easier, which is likely to trigger severe consequences in individuals’ social life. In this context, investigating detection algorithms is worthwhile, including approaches based on intrinsic noise inside different cameras, of image malicious manipulation.

Noise estimation lays crucial groundwork for noise-based blind image forensic algorithms. Most existing methods of noise estimation assume additive white Gaussian noise (AWGN), in which noise is represented by a fixed standard deviation given a single noisy image. In \cite{dong2017tip,zoran2009iccv,tang2015tcsvt,ma2020ietip}, researchers estimated the noise level by solving a nonlinear programming problem with respect to high-order image statistics, such as kurtosis and skewness. Instead of analyzing statistical properties, some approaches are patch-based. In \cite{pyatykh2013tip}, Pyatykh \textit{et al.} applied principle component analysis (PCA) to selected patches, estimating the noise intensity from the minimum eigenvalue of the covariance matrix. More recently, Wu \textit{et al.} \cite{wu2015eurasip-ijivp} estimated the noise level by exploiting irregular-shaped patches based on a superpixel scheme. However, AWGN conjecture does not hold for real-life digital photographs because actual CCD/CMOS sensor noise is strongly dependent on brightness. Thus, noise is better modeled as signal-dependent noise (SDN), where the noise standard deviation can be represented as a function of brightness. In \cite{foi2008tip}, Foi \textit{et al.} modeled the SDN as Poisson-Gaussian noise, where signal-dependent noise was represented by a Poisson model while the signal-independent component was demonstrated by a Gaussian model. Dong \textit{et al.} \cite{dong2018tip} proposed an effective SDN estimation method, in which regions with a frequently occurring intensity were selected to estimate the noise via constrained weight least-squares. Meanwhile, the investigators proposed in \cite{lin2004cvpr,lin2005cvpr,takamatsu2008cvpr,tai2013tpami,chen2019wcacv} that noise distribution for the radiance of each scene was likely to be skewed due to non-linear processing in the image sensor, which could be referred to as a camera response function (CRF). Accordingly, based on the observation of skewed noise, Liu \textit{et al.} \cite{liu2008tpami} modeled the sensor noise from the irradiance domain and converted the signal into the intensity domain using pre-measured CRF. This method assumed a piecewise smooth model, estimating noise by the way of Bayesian inference. This work was further extended in \cite{yang2012icip,yang20115tip}, where Yang \textit{et al.} introduced a sparse model for sensor noise, both in terms of estimation and denoising. Additionally, Thai \textit{et al.} \cite{thai2015sp} improved the Poisson-Gaussian model by employing gamma-correction to represent non-linearity.

Detecting and locating a splicing forgery in a digital image are usually based on the exposed significant variations of intrinsic characteristics that would otherwise be consistent in an untampered image. Noise or noise-related features can be used as a significant clue for distinguishing origins in a composited image due to its inevitable occurrence during the in-camera processing. Existing methods can be simply categorized into three classes: blind methods, prior information-needed methods, and deep learning-based methods.

Blind methods do not use any external data for training or other pre-processing, relying exclusively on the image itself to reveal the presence of manipulation. In \cite{mahdian2009ivc}, the noise variance was estimated by applying wavelet analysis to non-overlapping image blocks, and then a segmentation process was carried out to check for homogeneity. In \cite{lyu2014ijcv}, Lyu \textit{et al.} proposed a blind noise estimation approach to evaluate the noise level using projection kurtosis, and then designed a splicing detection method based on blind local variance estimation. Yao \textit{et al.} \cite{yao2018mta} also adopted the relation of projection kurtosis and noise to expose splicing forgery, incorporating an inhomogeneity scoring strategy to handle the impact of the complexity of the texture. Zeng \textit{et al.} \cite{zeng2017mta} utilized the K-means clustering method to classify original and suspicious regions according to the block noise variance obtained by a PCA-based method \cite{pyatykh2013tip}. Unlike the above approaches, which assumed a constant noise level imposed by the AWGN model across an untampered image, recent works have used an intensity-dependent noise model. In \cite{pun2016jvcir}, Pun \textit{et al.} proposed a Poisson model to indicate the primitiveness of noise artifacts based on noise level function (NLF) to reveal splicing. Yao \textit{et al.} \cite{yao2017mta} found inconsistency in noise level functions in different regions of a test aimed to locate splicing forgery. In \cite{zhu2018spic}, Zhu \textit{et al.} adopted an NLF to reflect the relationship between noise variance and the sharpness of block, generating a distance map for splicing detection. It is also worth mentioning that in \cite{liu2020neuc}, local noise variance was estimated by an adaptive SVD with an SVM training, which was exploited to find inconsistency. Another approach to exploit noise is the residual-based method. In \cite{cozzolino2015wifs}, the high-pass noise residuals of an image were exploited to extract rich features; next, the expectation maximization (EM) algorithm was then applied to reveal anomalies.

A photo-response non-uniformity (PRNU) noise pattern, which can be categorized as a prior information-needed method, has also been widely studied. Chen \textit{et al.} \cite{chen2008tifs} estimated the predictor of PRNU and noise residuals, then incorporated a decision test into a binary hypothesis problem. Korus \textit{et al.} \cite{korus2017tifs} proposed a PRNU-based tampering localization method using a multi-scale fusion approach. In turn, Cozzolino \textit{et al.} \cite{cozzolino2017iciap} proposed a blind localization method; however, it still needs an image dataset in advance for the clustering module. As an alternative, deep learning-based methods have commanded much attention in recent years. Barni \textit{et al.} \cite{barni2017jvcir} proposed a CNN-based approach, which worked on noise residuals to expose double JPEG compression traces. Bondi \textit{et al.} \cite{bondi2017cvpr} exploited CNN to extract characteristic camera model features from image patches and then localized the alien region employing iterative clustering techniques. More recently, Cozzolino \textit{et al.} \cite{cozzolino2020tifs} proposed a novel deep learning method to extract a noise residual, called noiseprint, for various forensic tasks, especially in image forgery localization. Generally, methods that require prior information and deep learning-based methods must meet the prerequisite that a large number of authentic images are available, which are known to come from the camera of interest, or image training datasets must be prepared in advance. However, such a scenario is not always reasonable in the real world, in which most detection scenarios are blind. Meanwhile, deep learning-based methods may be affected by any variation in the datasets during the test.

Since the noise level is dependent on image brightness, we used NLF to represent the noise characteristic of an image. We first proposed a conditional probability model using chi-square distribution, which was based on an approximate Gaussian model, to handle the different intensities of alien noise, considering non-linearity and signal-dependency at the same time. Meanwhile, physical properties in a neighborhood of image space were taken into account with a Markov random field (MRF), which could overcome the defects resulting from individual impact of noise intensity. We further inferred the forgery location map using the maximum \textit{a posterior} (MAP)-MRF framework, allowing the final decision map to be object-orientated and edge-smooth. Experiments were conducted to demonstrate that our approach outperformed other noise-based localization methods, yielding results that were both quantitatively convincing and visually pleasing. Moreover, our approach is distinctively automatic and blind to all images. The main highlights of this work can be summarized as follows:
\begin{enumerate}
\item We proposed a statistical assumption of signal-dependent noise, and analyzed the statistical characteristic of NLF, providing empirical evidence.
\item We introduced the MAP-MRF framework to blindly detect the splicing forgery with a conditional likelihood model between original noise and alien noise.
\item We solved the MRF parameters estimation via an iterative alternating strategy without any supervised training.
\end{enumerate}

The paper is organized as follows: Section \ref{statistical sdn} analyzes the statistical characteristics of NLF and provides empirical evidences. Section \ref{application} details the method for localizing a splicing forgery based on NLF and MAP-MRF, with experimental results shown in Section \ref{experiments}. Finally, Section \ref{conclusion} provides concluding remarks.

\section{Statistical Analysis of Signal-dependent Noise}\label{statistical sdn}
\subsection{Image noise modeling and analysis}
\begin{figure}[!t] 
	\centering
	\includegraphics[width=3.5in]{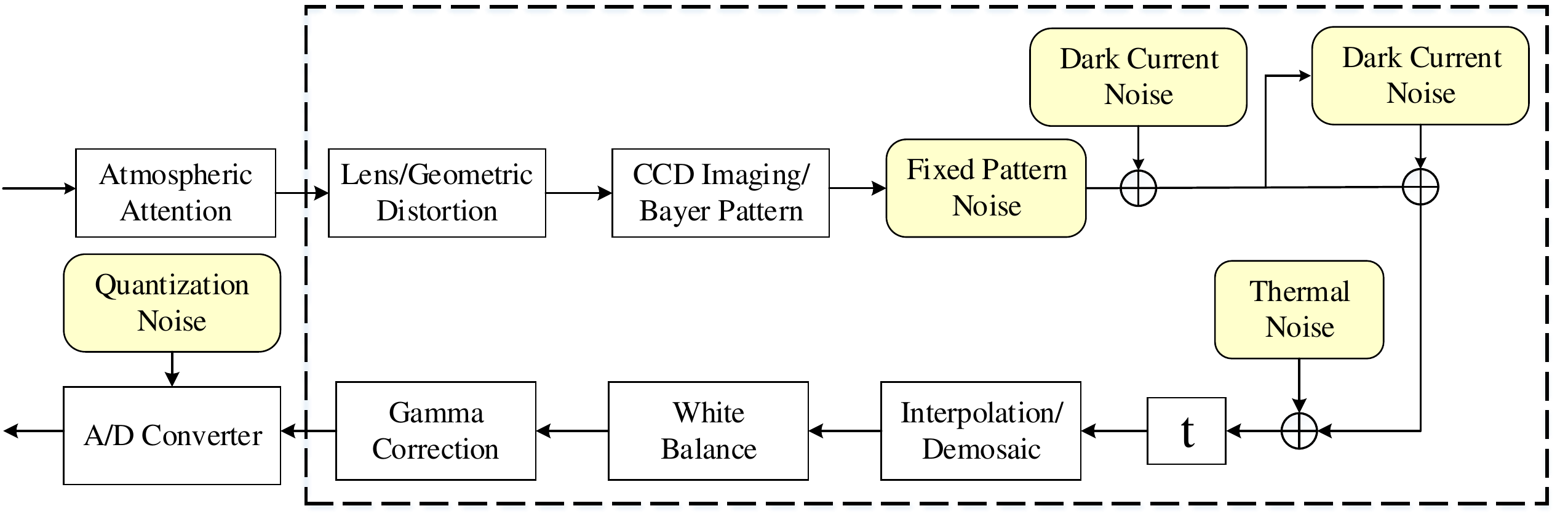}
	\caption{Imaging pipeline of CCD/CMOS camera, redrawn from \cite{liu2008tpami}.}
	\label{fig_1}
	\vspace{-0.5cm}
\end{figure}
Let $y\in\mathbb{R}$ be a digital image, defined on a rectangular lattice $\Omega $, with $y_i\in\Omega$, observed at the camera output, either as a single-color band or a composition of multiple color bands. Let us suppose, in a simplified model \cite{healey1994tpami}, that $y$ can be written as
\begin{equation}
\label{eq1}
y= x+n,
\end{equation}
where $x$ is the ideal noise-free image, while $n$ is the noise that accounts for all types of disturbance. Inside a camera, although the additive Gaussian noise model is widely assumed, noise exhibits signal-dependent behavior and greatly relies on the non-linear camera response function (CRF). According to \cite{liu2008tpami}, the noise model takes the form
\begin{equation}
\label{eq2}
\left\{
\begin{array}{lcl}
x=f(L),\\
y=f(L+n_s+n_c)+n_q,
\end{array}
\right.
\end{equation}
where $f\left ( \cdot  \right )$ is the CRF; $n_s$ denotes all the noise components that depend on irradiance $L$, that is, $n_s\sim\mathcal{N}(0,L\sigma_{s}^2)$; $n_c$ is the independent noise that is assumed $n_c\sim\mathcal{N}(0,\sigma_{c}^2)$; and $n_q$ is the additional quantization noise, which can be ignored in the model since it is quite small compared with other noise. Providing a better describe description of the noise characteristics inside a camera \cite{liu2008tpami}, the NLF is estimated as $\sigma \left ( x \right )= \sqrt{E\left ( y-x \right )^{2}}$. The estimation of NLF can be further rewritten as
\begin{equation}
\label{eq3}
\sigma \left ( x \right )=\sqrt{E\left ( f\left ( f^{-1}\left ( x \right ),f,\sigma _s,\sigma _c \right )-x \right )^{2}},
\end{equation}
where $\sigma_s$ and $\sigma_c$ are the standard deviation of $n_s$ and $n_c$, respectively.

Based on \cite{liu2008tpami}, we know that CRF dominates the shape of NLF; $\sigma_s$  and $\sigma_c$ dominate the numerical values in every intensity of NLF. In the following, we would like to study the influence of $\sigma_s$ and $\sigma_c$ in detail. Neglecting the CRF, noisy image $y$ can be modeled as the sum of $x$, $n_s$ and $n_c$, while $x$ is equivalent to irradiance $L$. Thus, the variance of $y$ can be derived as
\begin{equation}
\label{eq4}
\sigma^2(y)=\sigma^2(x)+\sigma^2(n_s)+\sigma^2(n_c)=x\sigma_{s}^{2}+\sigma_{c}^{2}.
\end{equation}
Therefore, noise statistics for $y$ can be expressed as $y\sim\mathcal{N}(x,x\sigma_s^2+\sigma_c^2)$, by which (\ref{eq1}) can also be rewritten as
\begin{equation}
\label{eq5}
y\approx x+\sqrt{x\sigma _{s}^{2}+\sigma _{c}^{2}}\times \xi _{n},
\end{equation}
where $\xi _{n}$ stands for a variable with standard normal distribution, i.e., $\xi _{n}\sim\mathcal{N}(0,1)$. From (\ref{eq5}), the assumption that noise is additive Gaussian noise model is farfetched. Only when $x$ is very small, which means $x\sigma_s^2$ may be negligible, is the noise level dominated by $n_c$, and the noise model can be assumed to be approximately Gaussian. However, $x\sigma_s^2$ cannot be ignored when $x$ is large; therefore the noise level is determined by $n_s$ and $n_c$ simultaneously.

\subsection{Empirical evaluation for noise distribution of a certain intensity}
\begin{figure}[!t]
	\centering
	\subfloat[NLFs with different $\sigma_{s}^2$]{
		\includegraphics[width=1.4in]{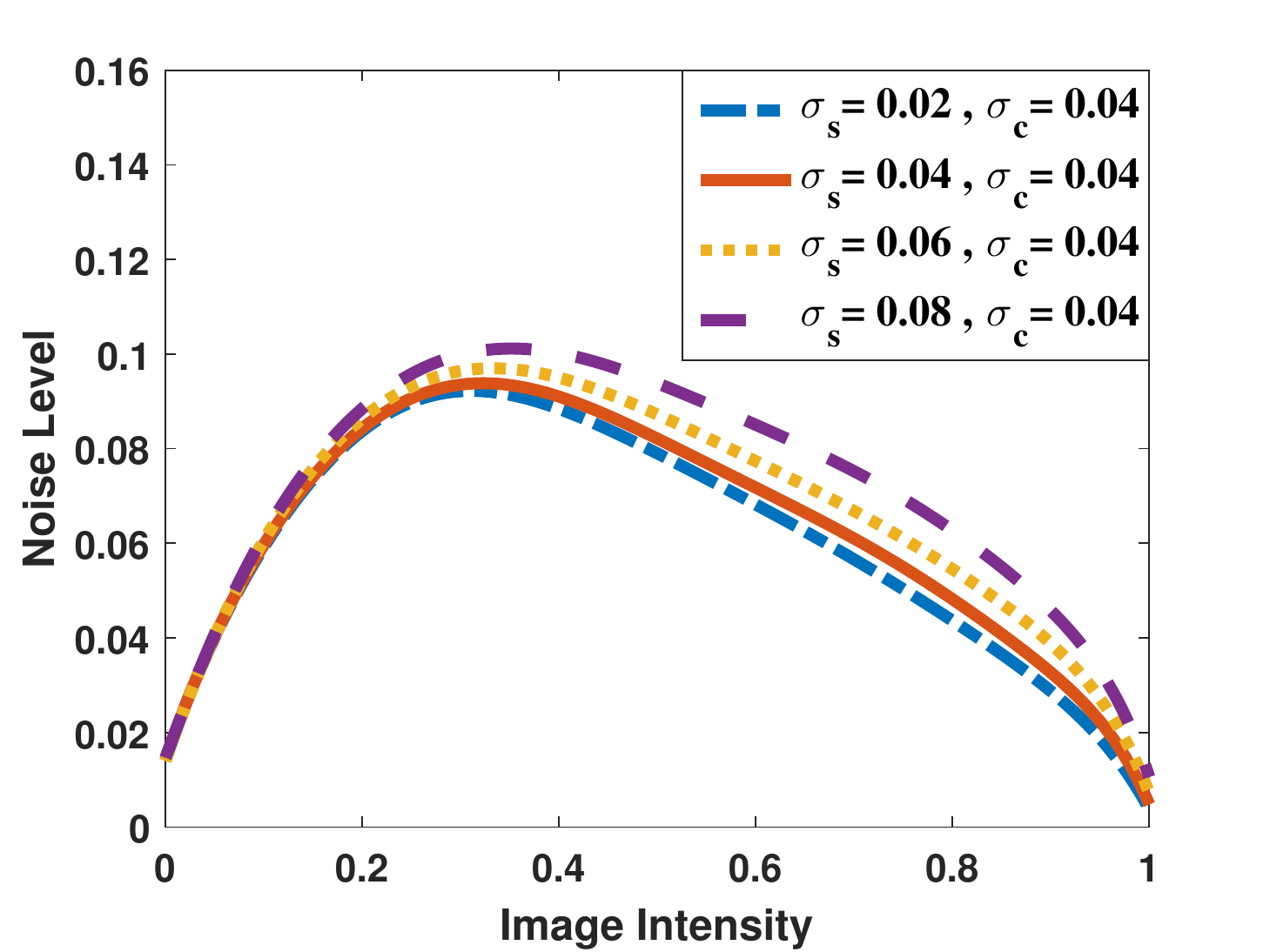}
		\includegraphics[width=1.4in]{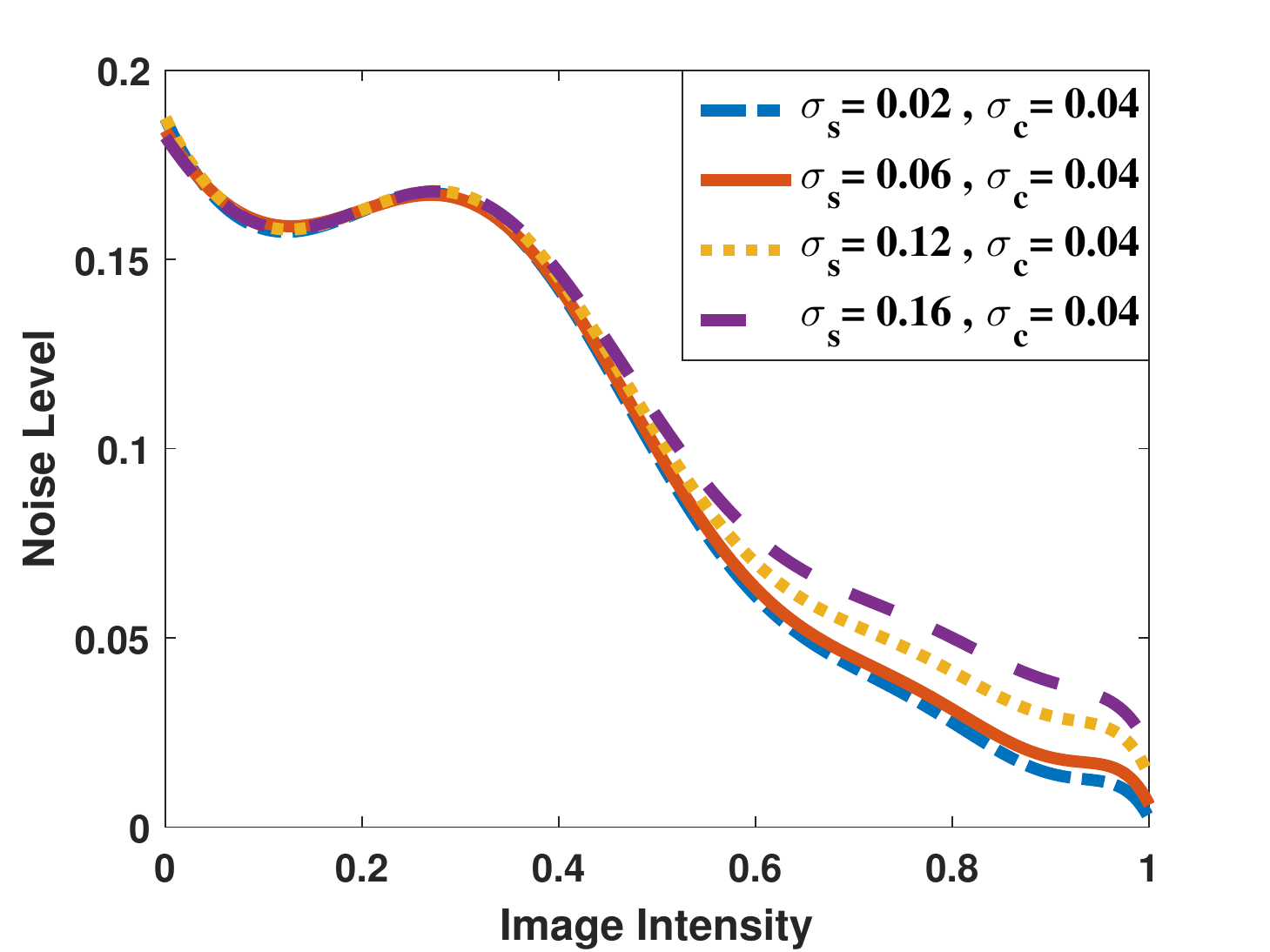}
	}
	\quad
	\subfloat[NLFs with different $\sigma_{c}^2$]{
		\includegraphics[width=1.4in]{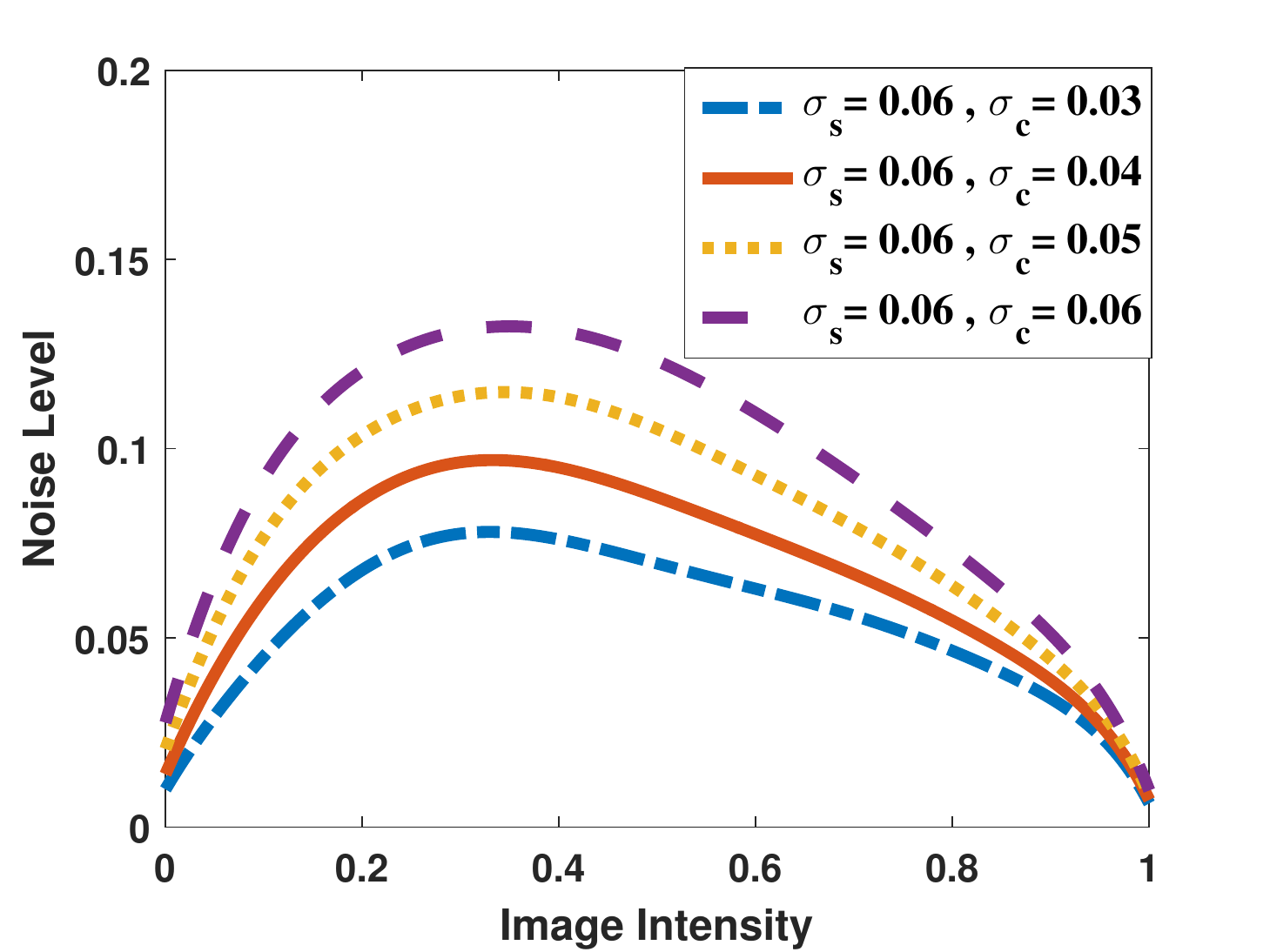}
		\includegraphics[width=1.4in]{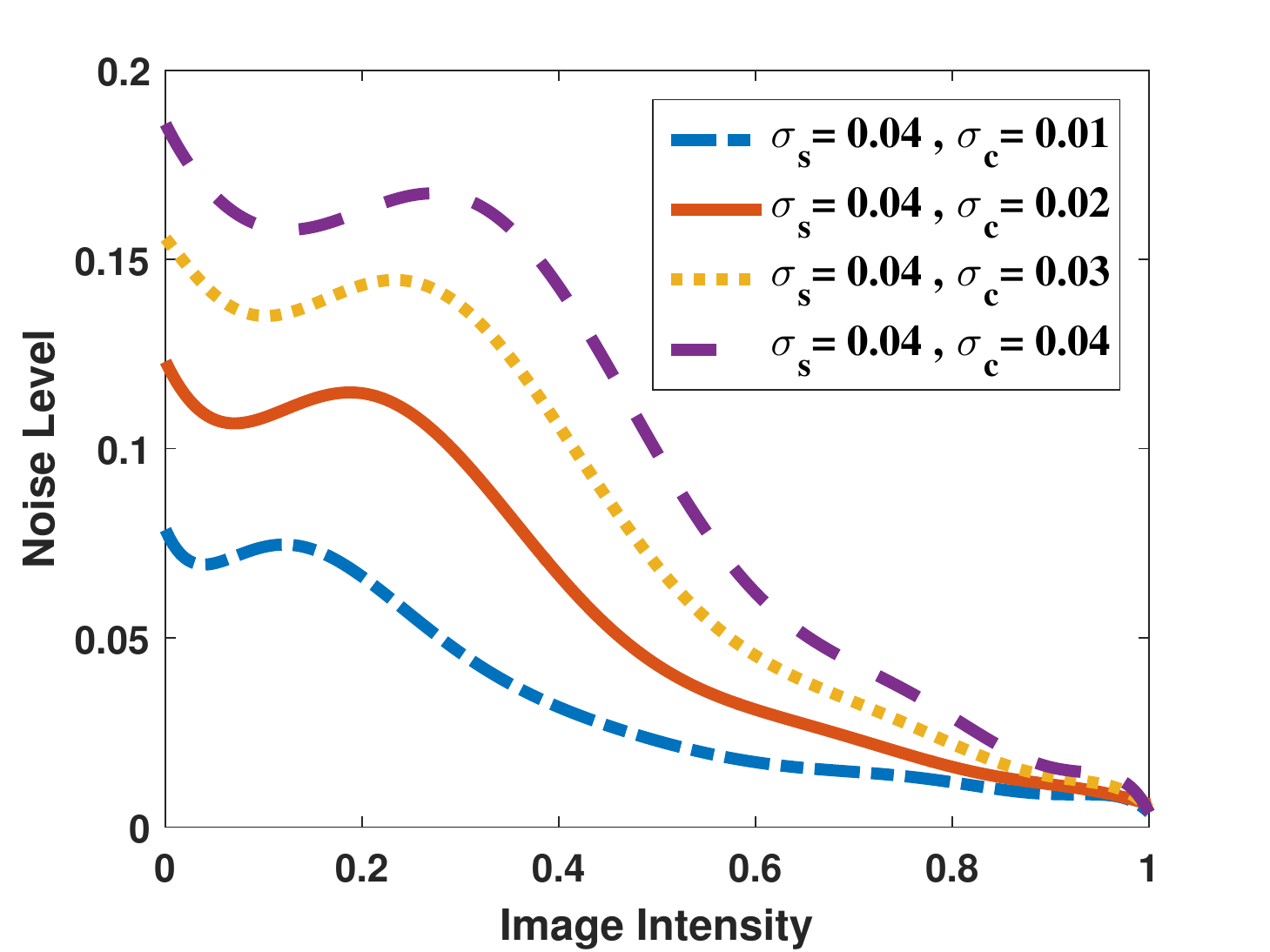}
	}
	\caption{Influence of change to NLF with different parameters. The synthesized NLFs in the left and right panels are derived from \emph{CRF-60} and \emph{CRF-50}, respectively. }
	\label{fig2}
	\vspace{-0.5cm}
\end{figure}
For better understanding, we chose two CRFs, \emph{CRF-50} and \emph{CRF-60}, as examples, both of which come from the Database of Response Functions (DoRF) from the CAVE Laboratory at Columbia University \cite{grossberg2004tpami}, to synthesize noise with different noise parameters according to Fig. \ref{fig_1}. The left and right panels in Fig. \ref{fig2} display the NLF curves produced by \emph{CRF-60} and \emph{CRF-50}, respectively. On the one hand, Fig. \ref{fig2}(a) reveals, for a fixed $\sigma_c=0.04$, the NLF curves for different $\sigma_s$ almost coincide during the interval of low intensity owing to the constant $\sigma_c$. However, during the interval of high intensity, the NLF curves show sustained growth as $\sigma_s$ increases. On the other hand, Fig. \ref{fig2}(b) shows that the change in $\sigma_c$ may significantly affect the noise level across the entire brightness range due to the independence of $\sigma_c$ from the intensity of the scene. Based on the above observations, it is reasonable to conclude that the noise changes more in a high-intensity interval than in a low-intensity interval with an identical change in $\sigma_s^2$, and the variation of $\sigma_c^2$ affects the noise level uniformly across the entire luminance interval.
\begin{figure}[!t]
	\centering
	\subfloat[Test pattern]{\includegraphics[width=1in]{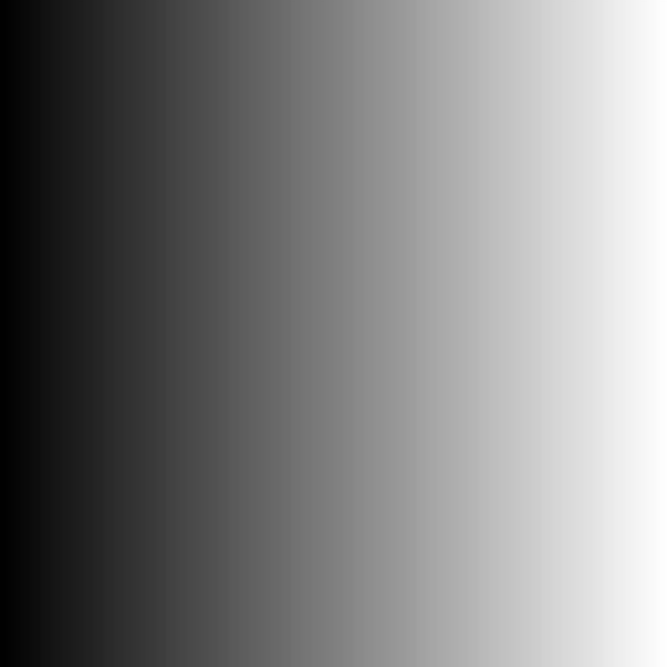}}\quad\quad\quad\quad
	\subfloat[Test pattern with added noise]{\includegraphics[width=1in]{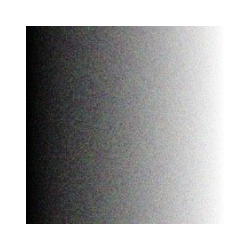}}
	\caption{An example of smoothly changing patterns. The test pattern of (a) is with a size of $1024\times1024$, where each column denotes an intensity level, and the noisy pattern in (b) is produced according to Fig. \ref{fig_1}.}
	\label{fig3}
	\vspace{-0.5cm}
\end{figure}
Eq. (\ref{eq5}) also demonstrates that real noise can be determined by an approximately Gaussian distribution form and transforms into an AWGN model under a single intensity value. Fig. 4 is a specific illustration for qualitative understanding. In Fig. 4, noise distributions in specific intensities are selected within regions close to the middle columns of the smoothly changing pattern shown in Fig. \ref{fig3}. Note that 6 typical camera response functions (\emph{CRF-27}, \emph{CRF-50}, \emph{CRF-53}, \emph{CRF-60}, \emph{CRF-100}, and \emph{CRF-155}), which are all saturated, are involved in the noise synthesis. Meanwhile, $\sigma_s$ and $\sigma_c$ were set at 0.06 and 0.04, respectively. Generally, Fig. 4 reveals that all distributions are approximately Gaussian-fitting, and some of which show a significantly high degree of fitting. In Fig. 4(a) and (b), the distributions of lower noise-free intensities are skewed to a certain extent, and these skewnesses mainly result from the convexity of \emph{CRF-27} and \emph{CRF-50}. However, these skewnesses are acceptable when considering only the general outline of the distributions.
\begin{figure}[!t]
	\centering
	\subfloat[]{\includegraphics[width=3.5in]{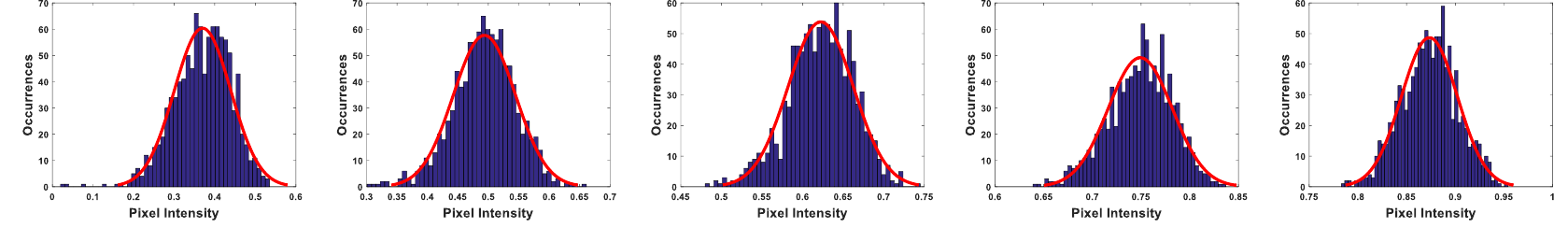}}
	\quad
	\subfloat[]{\includegraphics[width=3.5in]{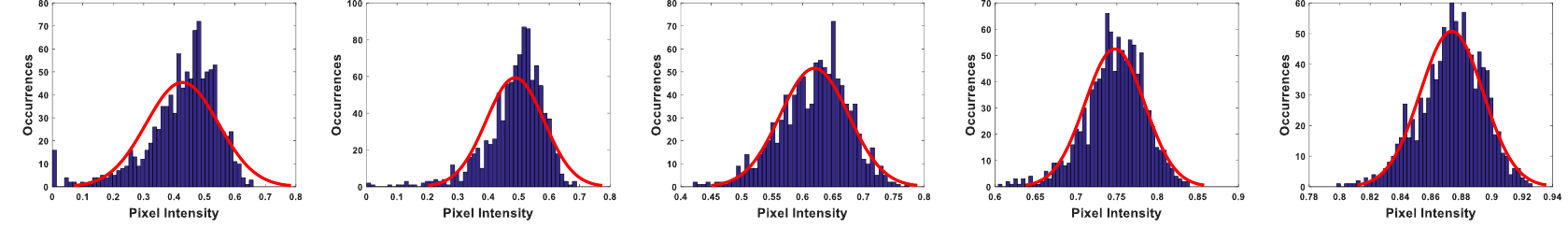}}
	\quad
	\subfloat[]{\includegraphics[width=3.5in]{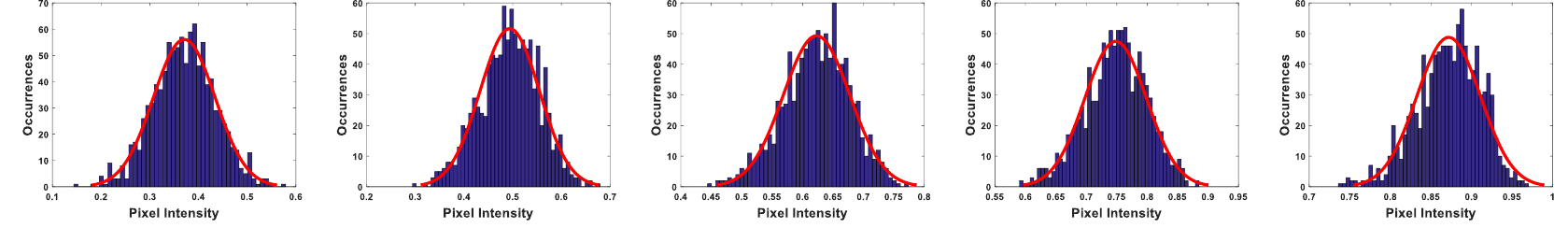}}
	\quad
	\subfloat[]{\includegraphics[width=3.5in]{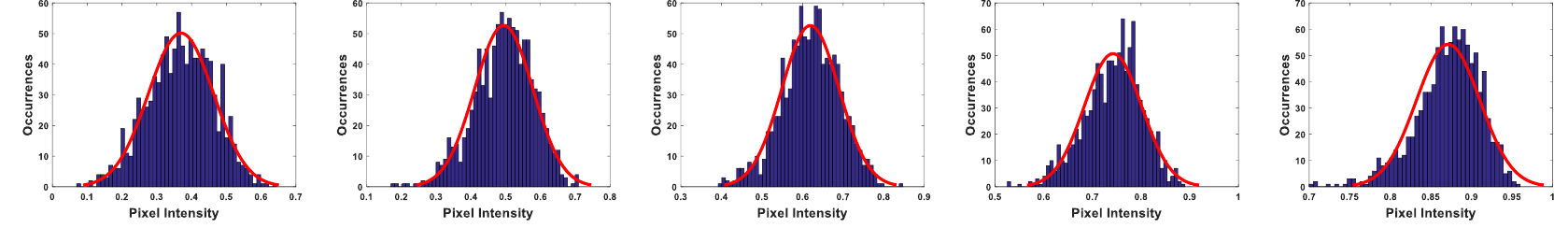}}
	\quad
	\subfloat[]{\includegraphics[width=3.5in]{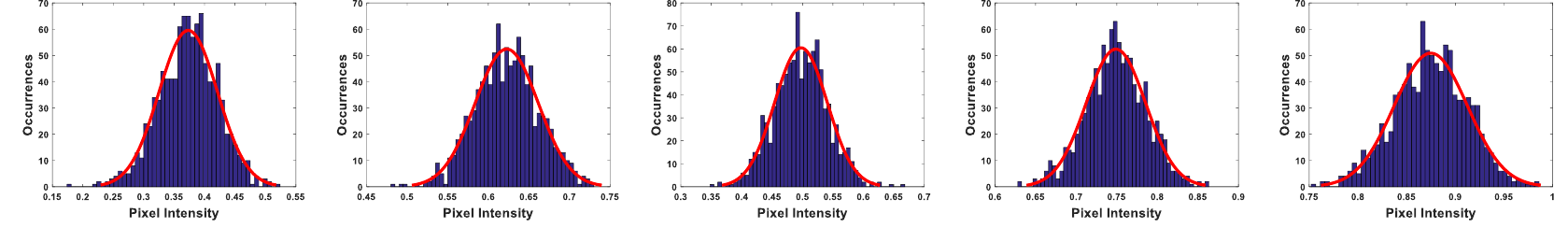}}
	\quad
	\subfloat[]{\includegraphics[width=3.5in]{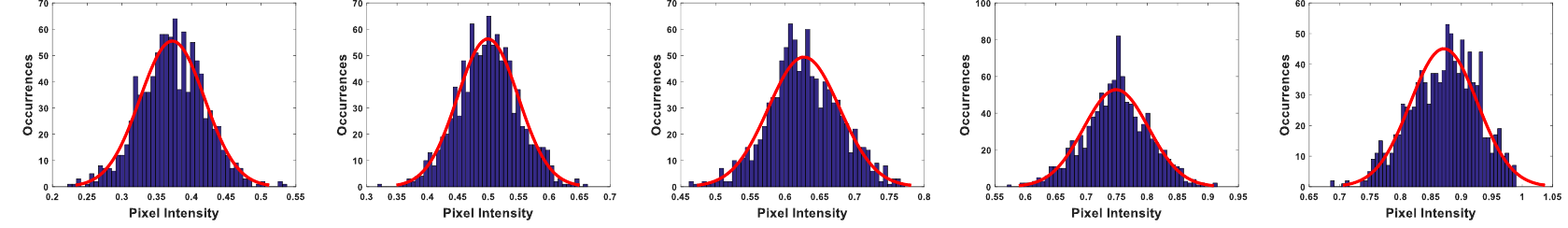}}
	\vspace{-0.1cm}
	\caption{Noise distribution and its corresponding fit to Gaussian distribution (red solid lines). From top to bottom, the distributions are derived from the noise added with \emph{CRF-27}, \emph{CRF-50}, \emph{CRF-53}, \emph{CRF-60}, \emph{CRF-100}, and \emph{CRF-155}, respectively. From left to right, noise-free intensities from the changing pattern are set at 0.375, 0.5, 0.625, 0.75, and 0.875, respectively.}
	\label{fig4}
	\vspace{-0.7cm}
\end{figure}

\begin{figure}[!t]
	\centering
	\subfloat[]{\includegraphics[width=3.5in]{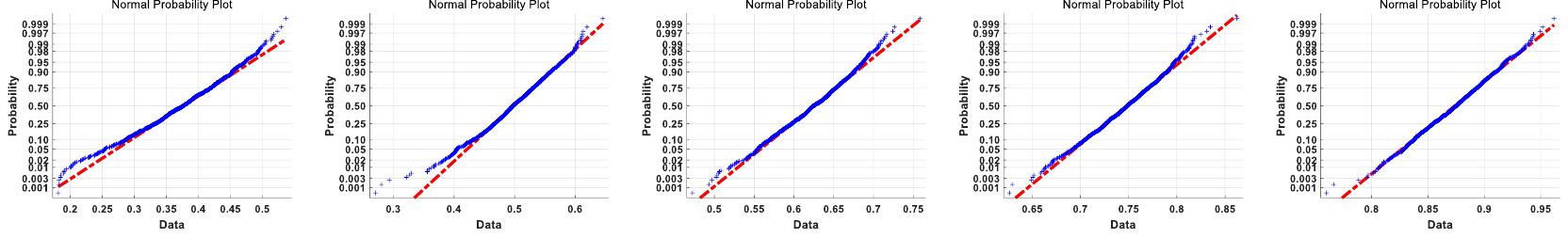}}
	\quad
	\subfloat[]{\includegraphics[width=3.5in]{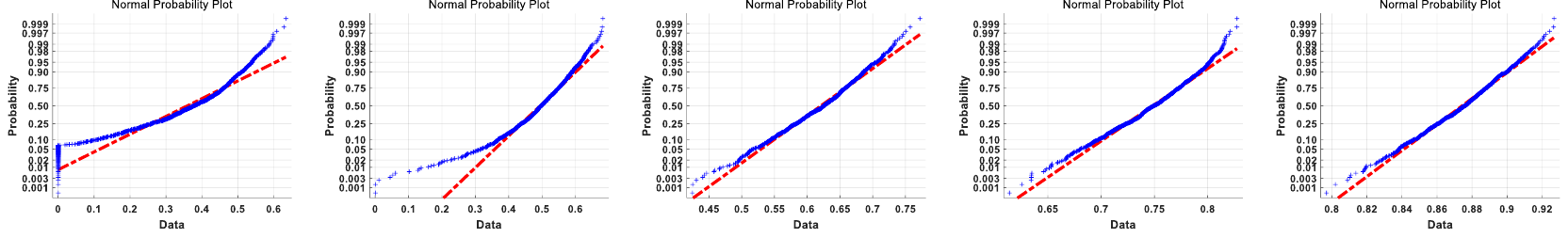}}
	\quad
	\subfloat[]{\includegraphics[width=3.5in]{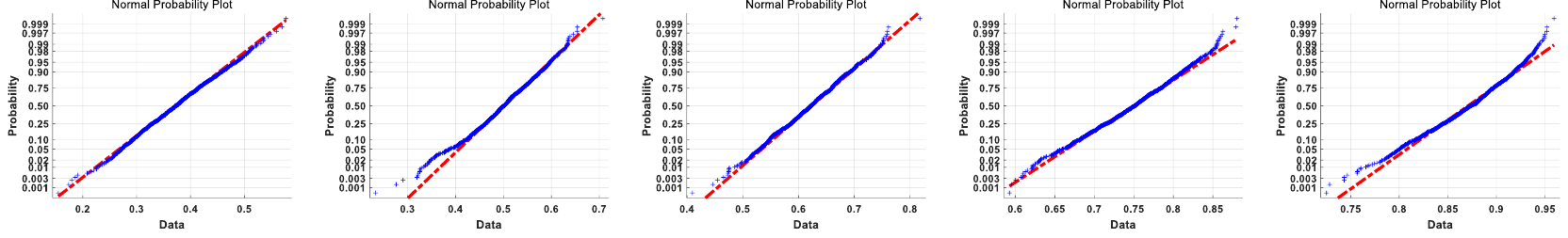}}
	\quad
	\subfloat[]{\includegraphics[width=3.5in]{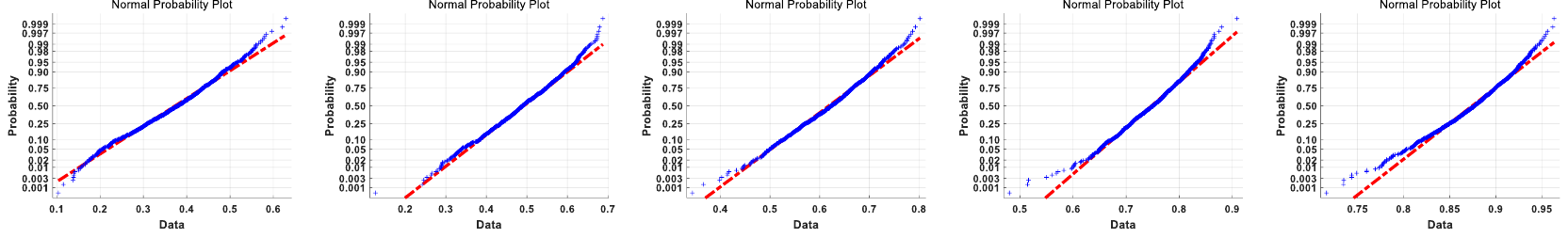}}
	\quad
	\subfloat[]{\includegraphics[width=3.5in]{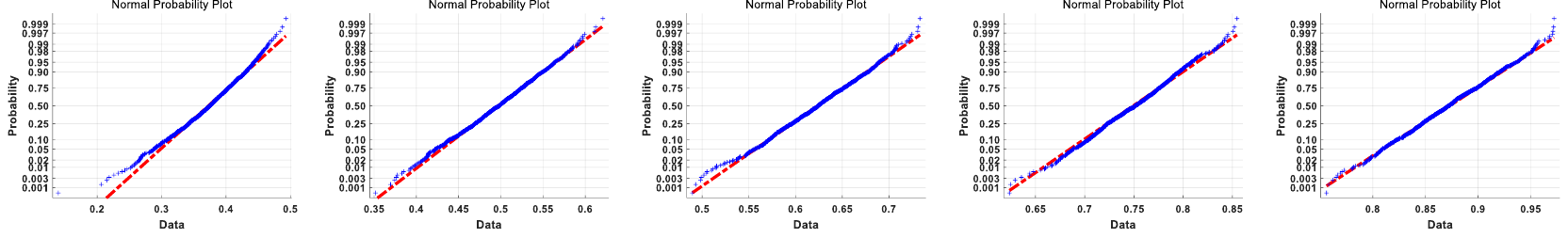}}
	\quad
	\subfloat[]{\includegraphics[width=3.5in]{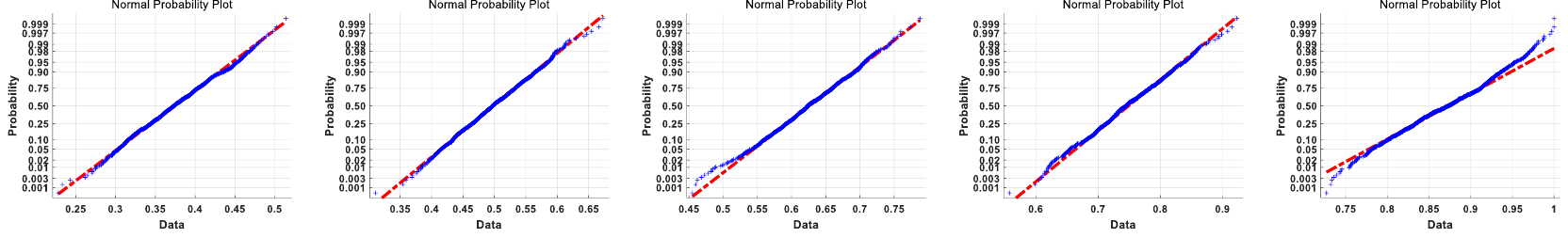}}
	\vspace{-0.1cm}
	\caption{Norm probability illustration, comparing the distribution of the noisy intensities to the normal distribution. From up to bottom, that is, (a)-(f), the distributions are derived from the noise added with \emph{CRF-27}, \emph{CRF-50}, \emph{CRF-53}, \emph{CRF-60}, \emph{CRF-100}, and \emph{CRF-155}, respectively. From the left to right, noise-free intensities from the changing pattern are set at 0.375, 0.5, 0.625, 0.75, and 0.875, respectively.}
	\label{fig5}
	\vspace{-0.5cm}
\end{figure}

To numerically evaluate the extent of approximation, we performed a regression analysis on all the samples of the selected noise-free intensities to provide a piece of empirical evidence. Fig. \ref{fig5} illustrates the norm probabilities, while Table I shows the numerical results. In Fig. \ref{fig5}, if the sample data has a Gaussian (or normal) distribution, then the data points appear along the reference line (denoted by a red dashed line). One can see that sample data fit the reference line well in (a), (c), (d), (e), and (f) of Fig. \ref{fig5}, and deviation points are few. In Fig. \ref{fig5}(b), a relatively large deviation exits between the data and curve, especially at both ends of the distribution interval. This phenomenon results from the non-linearity of CRF involved in noise production. However, as can be seen, the sample data concentrated at the noise-free intensity fit significantly well in Fig. \ref{fig5}(a)-(f), while Fig.4 indicates those data make up the majority of distributions. As a result, the deviation may be negligible to some extent. Meanwhile, Table \ref{table1} shows that root-mean-squared-error (RMSE) of the fit achieves a low level, which also generally suggests a small deviation. Another indicator, $R$-square ($R^2$) in Table I, demonstrates the goodness-of-fit, with a value closer to 1 indicating that a greater proportion of data is accounted for in the Gaussian distribution. Accordingly, Table I shows that each $R^2$ is very close to 1, and even $R^2$ of the noise distribution produced by \emph{CRF-50} with the largest fitting deviation yields a mean value of 0.9520.
\begin{table}[!t]
	\renewcommand{\arraystretch}{1.3}
	\renewcommand\tabcolsep{2.0pt} 
	\caption{Evaluating Goodness-of-fit in RMSE and $R$-square between benchmark probabilities and Gaussian-fitting probabilities.}
	\label{table1}
	\centering
	\begin{tabular}{cccccccc}
		\hline\hline
		Statistics & Intensity & \emph{CRF-27} & \emph{CRF-50} & \emph{CRF-53} & \emph{CRF-60} & \emph{CRF-100} & \emph{CRF-155}\\ \hline
		\multirow{6}*{RMSE}& 0.375& 0.1231& 0.2875& 0.0494 & 0.0876& 0.1244& 0.0510\\
		\cline{2-8}
		& 0.5 & 0.1470 & 0.3455 & 0.1419 & 0.0980 & 0.0547 & 0.0357\\
		\cline{2-8}
		& 0.625 & 0.0847 & 0.1084 & 0.0650 & 0.0829 & 0.0563 & 0.0587\\
		\cline{2-8}
		& 0.75 & 0.0787 & 0.1333 & 0.0828 & 0.1200 & 0.0758 & 0.0661\\
		\cline{2-8}
		& 0.875 & 0.0524 & 0.0916 & 0.1291 & 0.1507 & 0.0488 & 0.1312\\
		\cline{2-8}
		& mean & 0.0972& 0.1933& 0.0936& 0.1078& 0.0720& 0.0685\\ \hline\hline
		
		\multirow{6}*{$R^2$}& 0.375& 0.9848& 0.9173& 0.9976& 0.9923& 0.9845& 0.9974\\
		\cline{2-8}
		& 0.5& 0.9784& 0.8805& 0.9798& 0.9904& 0.9970& 0.9987\\
		\cline{2-8}
		& 0.625& 0.9928& 0.9882& 0.9958& 0.9931& 0.9968& 0.9966\\
		\cline{2-8}
		& 0.75& 0.8623& 0.9822& 0.9931& 0.9856& 0.9942& 0.9956 \\
		\cline{2-8}
		& 0.875& 0.9973& 0.9916& 0.9833& 0.9773& 0.9976& 0.9828\\
		\cline{2-8}
		& mean& 0.9631& 0.9520& 0.9899& 0.9877& 0.9940& 0.9942\\
		\hline\hline
	\end{tabular}
	\vspace{-0.3cm}
\end{table}

All the visual phenomenon and numerical results presented above suggest that noise distribution for a certain intensity can be fitted approximately by Gaussian distribution. Note that the noise model discussed here is limited to statistical similarity and does not involve a specific numerical calculation of noise. If specific noise values are to be computed, the effect of CRF should not be ignored.

\subsection{Conditional likelihood model of noise level function}
Since the distribution of signal-dependent noise can be gaussian approximately, its statistical property is easy to analyze if a group of static images with the same scene is available. However, in a real-life scenario, acquiring multiple images can be difficult; analysis often involves a single noisy image, in which only sample noise variance is obtained. Typically, noise variance of a certain intensity from a single image is obtained by calculating the local variance of an image patch in the spatial domain or transform domain. Estimated noise values represent sample variance, which might be close to a true population variance. Hence, we introduce chi-square value $\chi ^{2}$, defined as follows, to build a connection between sample variances and Gaussian distribution,
\begin{equation}
\label{eq6}
\chi ^{2}=\frac{\left ( n-1 \right )s^{2}}{\sigma ^{2}},
\end{equation}
where $s^2$ is sample variance, $\sigma^2$ is population variance, and $n$ is the number of obtained samples in a patch.\\
\textbf{Proposition 1.}$\,$ \emph{Suppose the number, denoted as $n$, of obtained samples in an image patch, the mean value, denoted as $m$, of the image patch and the NLF, denoted as $\sigma(x)$, are known; then the conditional likelihood of noise sample variance $s^2$ is derived as}
\begin{equation}
\label{eq7}
p\left ( s^{2}|\sigma ^{2} \right )\approx \chi _{n-1}^{2}\left ( \frac{\left ( n-1 \right )s^{2}}{\sigma ^{2}\left ( m \right )} \right ).
\end{equation}
\textbf{Proof.}$\,$ According to the statistical theory \cite{evans1993}, $\chi ^{2}$ is provided by a chi-square distribution with $n-1$ degrees of freedom, that is, $\chi ^{2}\sim\chi_{n-1}^2$, and the chi-square distribution yields us the likelihood of the sample variance $s^2$ under the assumption that population variance is $\sigma^2$. Thus, we can calculate the likelihood of $s^2$ from (\ref{eq6}), that is,
\begin{equation}
\label{eq8}
p\left ( s^{2}|\sigma ^{2} \right )= \chi _{n-1}^{2}\left ( \frac{\left ( n-1 \right )s^{2}}{\sigma ^{2}} \right ).
\end{equation}
Recalling that NLF as described by (\ref{eq3}) can represent the population variances, (\ref{eq7}) can be rewritten as follows:
\begin{equation}
\label{eq9}
p\left ( s^{2}|\sigma ^{2} \right )\approx p\left ( s^{2}|\sigma ^{2}\left ( m \right ) \right )=\chi _{n-1}^{2}\left ( \frac{\left ( n-1 \right )s^{2}}{\sigma ^{2}\left ( m \right )} \right ).
\end{equation}
Proposition 1 represents the likelihood of the sample variance $s^2$ conditioned by NLF $\sigma(m)$ upon obtaining a sample mean $m$. Using Proposition 1, we can further determine whether a region originates from the host image by the extension of (\ref{eq7}), which will be discussed in detail in Sec. \ref{application}.

\section{Application in Image Splicing Localization}\label{application}
In this section, we present our blind detection algorithm, which relies on NLF. Differing from some previous works based on noise inconsistency and blind circumstance, we transform the detection problem into a labeling task with a Markovian prior and deliver the conditional likelihood model of NLF into the optimization formulation, with which the labeling task will be solved. More details are introduced in the following subsections. Fig. 6 shows the flow diagram of the proposed method.
\begin{figure}[!t] 
	\centering
	\includegraphics[width=3.5in]{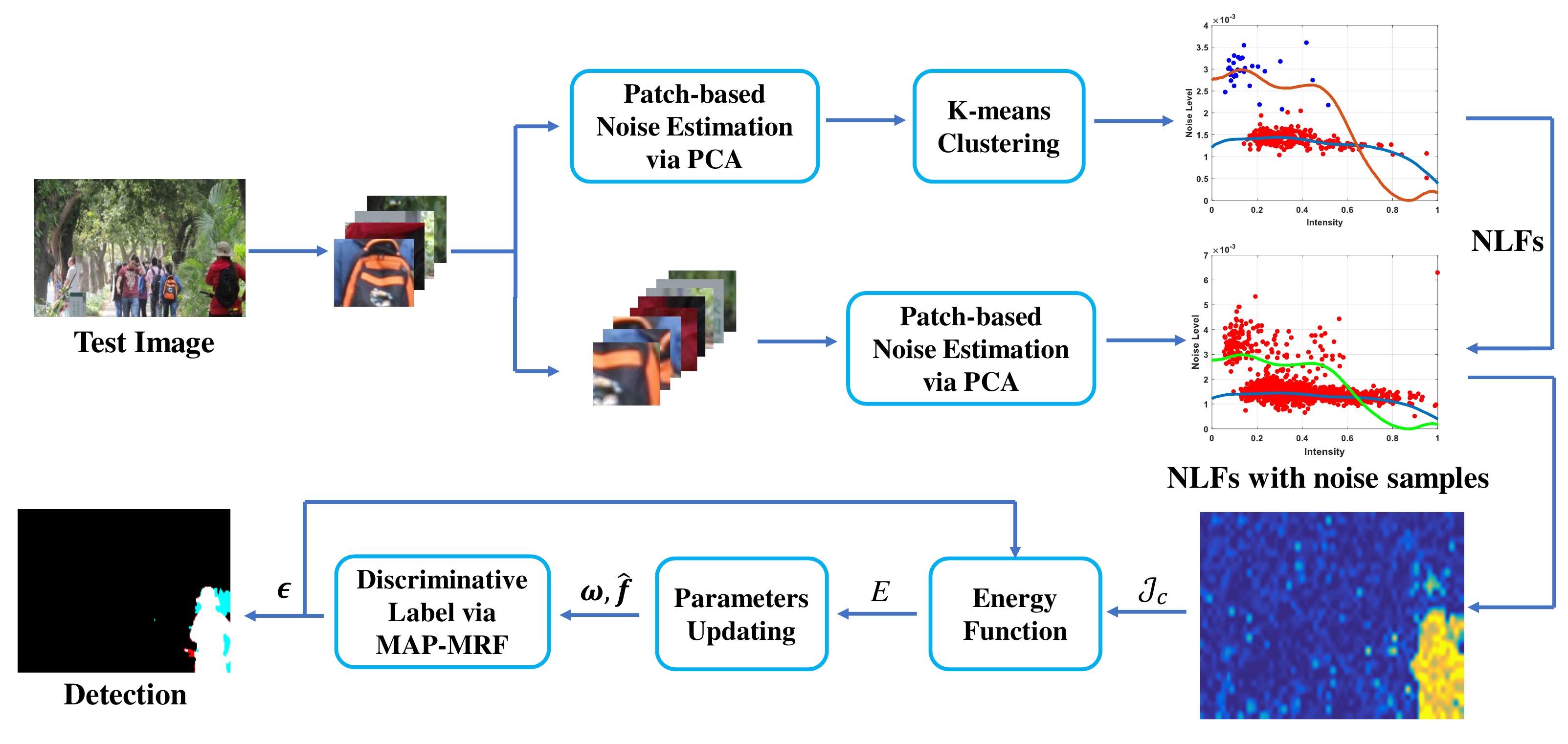}
	\caption{Block diagram of the proposed algorithm.}
	\label{fig_6}
	\vspace{-0.5cm}
\end{figure}
\subsection{Labeling problem formulation}\label{iii-a}
Our goal is to find a forensic map $\bm{f}$, which is binary and has the maximum probability to occur, given the observed data $\bm{d}$ from a single image,
\begin{equation}
\label{eq10}
\hat{\bm{f}}=\mathop{\arg\max}_{\bm{f}\in\left \{ 0,1 \right \}^{\mathit{N}}}p\left ( \bm{f}|\bm{d} \right ),
\end{equation}
where the value of estimated map $\hat{\bm{f}}$ labeling 1 means a forgery, whereas the 0 means the authentic part in the host image. Furthermore, we can rewrite the conditional distribution of (\ref{eq10}) as
\begin{equation}
\label{eq11}
\hat{\bm{f}}=\mathop{\arg\max}_{\bm{f}\in\left \{ 0,1 \right \}^{\mathit{N}}}\prod_{i=1}^{I}p\left ( d_i|f_i \right ) \prod_{i=1}^{I}p\left ( f_i \right ),
\end{equation}
where $I$ denotes the total number of pixels in the observed image. The first term of (\ref{eq11}) is a likelihood probability of $d_i$ conditional on $f_i$ and the second term accounts for the prior probability of labels guiding the forgery decision process towards reasonable results. In the following part, we will consider the Markov random field (MRF), in which the spatial or contextual dependencies exhibited by natural images can be taken into account, reflecting the prior distribution of labels. In MRF theory, an equivalence between MRF and Gibbs distribution is proved. Hence, the prior distribution takes the form \cite{li2001}:
\begin{equation}
\label{eq12}
p\left ( f \right )=Z^{-1}\times exp\left ( -\frac{1}{T}U\left ( f \right ) \right ),
\end{equation}
where $Z$ is a normalizing constant, $T$ is a constant called \emph{temperature}, which shall be assumed to be 1 unless otherwise stated, and $U(f)$ is the \emph{energy function}. The energy
\begin{equation}
\label{eq13}
U\left ( f \right )=\sum_{c\in \mathcal{C}}^{}V_{c}\left ( f \right )
\end{equation}
is a sum of \emph{clique potentials} $V_c (f)$ over all possible cliques $\mathcal{C}$. The value of $V_c (f)$ depends on the local configuration of clique $c$. A clique $c$ for $(\mathcal{S},\mathcal{N})$ is defined as a subset of sites in $\mathcal{S}$, where $\mathcal{S}$ contains the nodes $i$, and $\mathcal{N}$ determines the links between the nodes according to the neighboring relationship. Considering the collection $\mathcal{C}$ of all cliques for $(\mathcal{S},\mathcal{N})$, (\ref{eq13}) can be expanded as the sum of several terms as follows:
\begin{equation}
\label{eq14}
\begin{aligned}
U\left ( f \right )=&\sum_{\{i\}\in\mathcal{C}_1}^{}V_1\left ( f_i \right )+\sum_{\{i,i'\}\in\mathcal{C}_2}^{}V_2\left ( f_i,f_{i'} \right  )\\
+&\sum_{\{i,i',i''\}\in\mathcal{C}_3}^{}V_3\left ( f_i,f_{i'},f_{i''} \right )+\cdots .
\end{aligned}
\end{equation}
In most cases, contextual constraints on both labels are widely used because of their simple form and low cost in computation, and they are encoded in the Gibbs energy as pair-site clique potentials. With clique potentials of up to 2 sites, the energy takes the form
\begin{equation}
\label{eq15}
U\left ( f \right )=\sum_{i\in\mathcal{S}}^{}V_1\left ( f_i \right )+\sum_{i\in\mathcal{S}}^{}\sum_{i'\in\mathcal{N}}^{}V_2\left ( f_i,f_{i'} \right ),
\end{equation}
where “$\sum_{i\in\mathcal{S}}^{}$” is equivalent to “$\sum_{\{i\}\in\mathcal{C}_1}^{}$” and “$\sum_{i\in\mathcal{S}}^{}\sum_{i'\in\mathcal{N}}^{}$” equivalent to “$\sum_{\{i,i'\}\in\mathcal{C}_2}^{}$”. A particular MRF can be specified by properly selecting $V_1$ and $V_2$. For single-site cliques, the clique potentials depend on the label as
\begin{equation}
\label{eq16}
V_1\left ( f_i \right )=\alpha,\, \,{\rm if}\ \, f_i=1,
\end{equation}
where $\alpha$ is the penalty against tampered authentication units. The higher $\alpha$ is, the fewer pixels will be assigned a value of 1. This outcome has the effect of controlling the percentage of sites labeled 1. As for pairwise contextual constraints, it assumes that physical properties in a neighborhood of space present some coherence and generally do not change abruptly. In a simple case, the pair-site clique potentials can be defined as
\begin{equation}
\label{eq17}
V_2\left ( f_i,f_{i'} \right  )=g\left ( f_i-f_{i'} \right ),
\end{equation}
where $g\left ( f_i-f_{i'} \right )$ is a function penalizing the violation of smoothness caused by the difference $f_i-f_{i'}$. Here, we resort to a function taking the form
\begin{equation}
\label{eq18}
g\left ( f_i-f_{i'} \right )=\beta _{i,i'}\left | f_i-f_{i'}  \right |,
\end{equation}
where $\beta_{i,i'}$ is the penalty against nonequal labels on two-site cliques. This value can be calculated as follows \cite{korus2017tifs}:
\begin{equation}
\label{eq19}
\beta _{i,i'}=\beta _{0}+\beta _{1}e^{-\frac{1}{2}\phi^{-2}\left \| d_i,d_{i'} \right \|_{2}^{2}},
\end{equation}
where $\left \| d_i,d_{i'} \right \|_{2}^{2}$ denotes L2 distance between two pixels (computed from RGB vectors). The first term encodes a default, content-dependent penalty. The second term represents the interactions of neighboring authentication units, with similarity attenuation controlled by parameter $\phi$ (empirically chosen to be 25). Thus, (\ref{eq18}) guides a preference towards piecewise-constant- and contextual-dependency-based solutions.

Hence, we can rewrite (\ref{eq11}) by taking the negative log and combining all the prior models above, as follows:
\begin{equation}
\label{eq20}
\begin{aligned}
\hat{\bm{f}}=\mathop{\arg\min}_{\bm{f}\in\left \{ 0,1 \right \}^{\mathit{N}}}&\left \{ \sum_{i=1}^{I}-log\,p\left ( d_i|f_i \right )+\alpha \sum_{i\in\mathcal{S}}^{}f_i \right.\\
&\left.  +\beta_{i,i'}\sum_{i\in\mathcal{S}}^{}\sum_{i'\in\mathcal{N}}^{}\left | f_i-f_{i'}  \right |\right \}.
\end{aligned}
\end{equation}
The second and third terms are the penalty terms and have been discussed above. The first term (referred to as the \emph{likelihood energy}) depends only on the data.

\subsection{Image tamper probability derivation}\label{iii-b}
Since $log\,p\left ( d_i|f_i \right )$ in (\ref{eq20}) is a likelihood of observation $d_i$ conditioned on label $f_i$, we will discuss here how the image tamper probability mixed with a different label $f_i$ determines the likelihood energy. When a tamper probability $P_{t_i}$ for a patch is acquired, the label $f_i$ assigned to it should be consistent with the requirement of optimization (\ref{eq20}). That is, the wrong label leads to maximum likelihood energy, and the right one leads to minimum energy. Based on this principle, we simply apply the following equation:
\begin{equation}
\label{eq21}
log\,p\left ( d|f \right )=\left\{\begin{matrix}
	\begin{aligned}
		&log\,\left(1-P_{t_i}\right)&,\,\,{\rm for} f=0,\\
		&log\,P_{t_i}&,\,\,{\rm for} f=1.
	\end{aligned}
\end{matrix}\right.
\end{equation}
A smaller $P_{t_i}$ indicates an authentic observation, and if $f=1$, which means a wrong inferred label, the $log\,p\left ( d_i|f_i \right )$ will be small while energy is opposite due to its taking a negative value in (\ref{eq20}); and vice versa.

Assuming that two different NLFs have previously been obtained precisely from an image, we are ready to derive the tamper probability. Let one NLF, referred to as $\sigma_0$, corresponds to the pristine, and the other NLF, referred to as $\sigma_1$, corresponds to tampering. Hence, the likelihood of tampering can be derived, according to Bayes’ theorem and the law of total probability \cite{papoulis1984}, as follows:
\begin{equation}
\label{eq22}
\begin{aligned}
P_t&=p\left ( s^{2}\in \sigma _{1}\left ( m \right )|s^{2} \right )\\
&=\frac{p\left ( s^{2} |s^{2}\in\sigma _{1}\left ( m \right ) \right )p\left ( s^{2}\in\sigma _{1}\left ( m \right )  \right )}{\sum_{k=0}^{1}p\left ( s^{2} |s^{2}\in\sigma _{k}\left ( m \right ) \right )p\left ( s^{2}\in\sigma _{k}\left ( m \right )  \right )},
\end{aligned}
\end{equation}
where $p\left ( s^{2} |s^{2}\in\sigma _{1}\left ( m \right ) \right )$ can be easily calculated by (\ref{eq8}), and $p\left ( s^{2}\in\sigma _{1}\left ( m \right )  \right )$ takes the form:
\begin{equation}
\label{eq23}
p\left ( s^{2}\in\sigma _{1}\left ( m \right )  \right )=\frac{\sum_{i}^{}p\left ( s^{2}_{i} |s^{2}_{i}\in\sigma _{1}\left ( m \right ) \right )}{\sum_{k=0}^{1}\sum_{i}^{}p\left ( s^{2}_{i} |s^{2}_{i}\in\sigma _{1}\left ( m \right ) \right )}.
\end{equation}
Note that (\ref{eq22}) is accurate when the assumption that two NLFs can be obtained precisely is satisfied. Generally, an NLF is constructed by the sample variances in an image, and its precision depends on the number and estimation accuracy of noise samples. The former is determined by the image itself, while the latter depends on different methods.
\vspace{-0.3cm}
\subsection{Construction of NLFs in a tampered image}\label{iii-c}
As the estimation of NLF is not our main work, we have followed \cite{liu2008tpami} to construct the NLF. However, because the estimation of noise in \cite{liu2008tpami} is slightly overestimated, we have resorted to a PCA-based method \cite{pyatykh2013tip} to estimate noise variances.

First, the test image is decomposed into $N$ non-overlapping $B \times B$ patches. Next, we estimate noise variance $s_n^2$ in every single patch by a PCA-based method. Meanwhile, the mean value $m_n$ of every patch is calculated as the noise-free intensity. Then, we preliminarily and simply locate several suspicious regions, according to the numerical differences only of noise variances, by running the K-means clustering algorithm. Consequently, a noise sample set $\{\left(m_n,s_n \right)\}$ is decomposed into two sets, $\left \{ \left ( m_{n}^{[0]},s_{n}^{[0]} \right ) \right \}$ and $\left \{ \left ( m_{n}^{[1]},s_{n}^{[1]} \right ) \right \}$, in which superscripts 0 and 1 represent the original and suspicious regions, respectively. After obtaining the noise sample set $\left \{ \left ( m_{n}^{[i]},s_{n}^{[i]} \right ) \right \}$, a Bayesian MAP inference is exploited to construct NLF $\sigma^{[i]}$, as follows:
\begin{equation}
\label{eq24}
\begin{aligned}
&\hat{x}_l=\mathop{\arg\min}_{x_l}\sum_{n}\left [ {\rm -log\,}\Phi\left ( \frac{\sqrt{k_n}}{s_n^{[i]}}\left ( s_n^{[i]}-e_n^Tx_l^T-\bar{\sigma}^{[i]}\left ( m_n^{[i]} \right ) \right ) \right ) \right.\\
&\left.+\frac{\left ( e_n^Tx_l^T+\bar{\sigma}^{[i]}\left ( m_n^{[i]} \right )-s_n^{[i]} \right )^{2}}{2\left ( s_n^{[i]}/\sqrt{k_n} \right )^{2}}\right ]+x_l^T\Lambda ^{-1}x_l,\\
\end{aligned}
\end{equation}
subject to
\begin{equation}
\label{eq24-1}
\sigma ^{[i]}=\bar{\sigma} ^{[i]}+\mathbf{E}x_l\geq 0.
\end{equation}
In the above formula, $k_n$ denotes the number of pixels in a patch; the matrix $\mathbf{E}=\left [ \omega_1,\cdots,\omega_m  \right ]\in \mathbb{R}^{d\times m}$, where $d=256$, contains the principal components; $e_n$ is the $n$-th row of $\mathbf{E}$; and $\Lambda ={\rm diag}(v_1,\cdots,v_m)$ is the diagonal eigenvalue matrix whose elements correspond to principal components $\omega_m$ after PCA.

Note that all the samples involved in the construction of NLF originate from smooth or flat patches. By doing so, some overestimated sample variances can be avoided that would otherwise affect NLF.

\subsection{Combination strategy for refinement of tamper probability}\label{iii-d}
Recall from Sec. \ref{iii-b} that (\ref{eq22}) regarding tamper probability is accurate when the assumption that two NLFs can be obtained precisely is strictly held. Generally, the pristine region of a tampered image represents the majority and the noise sample size in this region is also large enough, which is likely to result in an accurate NLF. However, when the size of the tampered region is quite small, the noise sample size will also consequently be small, which may directly lead to an inaccurate NLF estimation originating from the tampered region. In the following dicussion, we are ready to study the strategy of refining inaccuracies in tamper probability.

Considering that NLF estimated from the original area is relatively accurate, we regard it as a benchmark. Note that original area here is an area initially taken as pristine, and it is generally credible even if some of noise from the patch in the original area are overestimated or a few tampered regions are initially mistaken as the original. Inspiring by the distance-based method for tampered detection \cite{zhu2018spic}, we calculate the tampered probability as
\begin{equation}
\label{eq25}
P_t=1-e^{-\zeta \left \| s_i-\sigma^{[0]}\left ( m_i \right ) \right \|_{2}},
\end{equation}
where $\zeta$ is a difference-expanding operator (empirically chosen as 50), $s_i$ denotes noise variances from a $B/2\times B/2$ sized patch in an image, and $\sigma^{[0]}\left ( m_i \right )$ signifies the benchmark NLF value of the sample mean $m_i$. Thus, (\ref{eq25}) indicates the distance between the estimated noise sample and reference NLF representing the authentic baseline. The closer the noise sample is to the reference curve, the less the probability is. In other words, the region where the sample comes from is more likely to be authentic. Note that we are calculating the noise variance in a smaller patch with a size of $B/2\times B/2$, and applying them to (\ref{eq22}) and (\ref{eq25}). The purpose is to obtain more samples and make classification softer.

Now we face the problem of how to choose the proper likelihood. On the one hand, when the size of the tampered region is small, then we will have enough samples to construct the benchmark NLF, by which the likelihood based on the benchmark curve will be relatively reliable. On the other hand, if the size of the tampered region is large enough, (\ref{eq22}) for tamper probability may yield better results. Here, we make a tradeoff to create a combination, resorting to a “steep” logistic function of shifted coordinates, as follows:
\begin{equation}
\label{eq26}
\beta_{p}=\frac{1}{1+e^{-\lambda\left ( A_r-\delta \right )}},
\end{equation}
where $\beta_{p}$ is a coefficient assigned to the interaction likelihood to determine the percentage in the combination, $A_r$ denotes the size of the tampered region, $\lambda$ is the strength of the steepness, and $\delta $ represents how much the coordinates shift. Consequently, we can compute the combination $\mathcal{J}_c$ as
\begin{equation}
\label{eq27}
\mathcal{J}_c=\left ( 1-\beta_p \right )\mathcal{J}_1+\beta_p\mathcal{J}_2,
\end{equation}
where $\mathcal{J}_1$ denotes the likelihood based on the benchmark curve, as listed in (\ref{eq25}), and $\mathcal{J}_2$ indicates the likelihood based on (\ref{eq22}). For further reference, Appendix \ref{appendix} provides a concrete numerical analysis between two probabilities $\mathcal{J}_1$ and $\mathcal{J}_2$.

\subsection{Parameter estimation with a self-iterative strategy in MAP-MRF}\label{iii-e}
Recalling the formulation of (\ref{eq20}), we have three unknown parameters $\bm{\omega} =\left \{ \alpha ,\beta _0,\beta _1 \right \}$, which are vital for the final solution in MRF framework. Traditionally, the selection of parameters is a supervised training process on the fully labeled training image examples. Such an operation relies on the assumption that a large number of images is available, which are known to the tampered region of interest. However, in a blind scenario, we can only assume that we have a certain number of images, whose origin is unknown. Inspired by \cite{takamatsu2008cvpr}, we will use an iterative alternating algorithm to estimate the parameters. The algorithm iterates between two steps: parameter estimation (looking for optimal $\bm{\omega}$) and inference (looking for optimal $\bm{f}$ based on estimated $\bm{\omega}$ of the current step). Hence, we will formulate the estimation problem in an energy minimization framework as
\begin{equation}
\label{eq28}
\left( \hat{\bm{f}},\,\hat{\bm{\omega}} \right)=\mathop{\arg\min}_{\bm{f},\,\bm{\omega}}E\left (\mathcal{J}_c,\bm{\omega},\bm{f} \right ),
\end{equation}
where the energy function is equivalent to the right-hand side of (\ref{eq20}). Algorithm I presents a detailed demonstration.
\begin{algorithm}[!t]
	\caption{Parameter estimation using an iterative alternating algorithm}
	\begin{algorithmic}[1]
		\REQUIRE  The combination of likelihoods $\mathcal{J}_c$ and the maximum iteration \emph{Iter\_max}
		\ENSURE $\bm{\omega} =\left \{ \alpha ,\beta _0,\beta _1 \right \}$ and final decision label $\hat{\bm{f}}$
		\STATE Initialize the label $\bm{f}$ approximately by K-means method.
		\REPEAT
		\STATE Minimize the energy $E$ with fixed input label $\bm{f}$ to estimate the $\bm{\omega}$ with $\hat{\bm{\omega }}=\mathop{\arg\min}_{\bm{\omega }}E$.
		\STATE Update $\bm{\omega}$: $\bm{\omega}\leftarrow{\hat{\bm{\omega}}}$.
		\STATE Fix the MRF parameter $\bm{\omega}$ and minimize the energy $E$ according to $\hat{\bm{f}}=\mathop{\arg\min}_{\bm{f}}E$.
		\STATE Update $\bm{f}$: $\bm{f}\leftarrow{\hat{\bm{f}}}$.
		\UNTIL $E$ converges or \emph{Iter\_max} reached.
	\end{algorithmic}
	\label{algorithm_1}
\end{algorithm}

We will construct MRF using the UGM toolbox \cite{schmidt} for undirected graphical models and a max-flow algorithm for faster GraphCuts \cite{boykov2004tpami}. Owing to higher accuracy and better convergence of LBP inference in the UGM toolbox, we have observed that most test cases converged before reaching maximum iteration $Iter\_max$ (we set it at 5 for rapidity). Even some cases reach the $Iter\_max$ without convergence, the final decision $\hat{\bm{f}}$ can also offer relatively good accuracy. Note the above inference process is completely unsupervised and no training sets is necessary. Given a new test image, the optimal parameters and the inferred labels are estimated.

The complete implementation of the forgery localization is presented in Algorithm II.
\begin{algorithm}[!t]
	\caption{The coefficients estimation with respect to the NLF from a single image}
	\begin{algorithmic}[1]
		\REQUIRE  Splicing tampered image $\bm{y}$
		\ENSURE Decision map of forgery localization $\hat{\bm{f}}$
		\STATE Decompose the input image $\bm{y}$ into $N$ non-overlapping blocks with a size of $B\times B$.
		\STATE Estimate the noise variance in each block using PCA-based method, and noise sample set $ \left \{ \left ( m_n,s_n \right )|n\in\left \{ 1,2,\cdots,N \right \} \right \}$ is obtained.
		\STATE Use K-means clustering algorithm to preliminarily and simply decompose noise sample set into $\left \{ \left ( m_n^{[0]},s_n^{[0]} \right ) \right \} $ and $\left \{ \left ( m_n^{[1]},s_n^{[1]} \right ) \right \} $.
		\STATE Select noise samples originated from smooth blocks to construct the NLFs, $\sigma^{[0]}$ and $\sigma^{[1]}$, by using (\ref{eq24}).
		\STATE Decompose the input image $\bm{y}$ into $4N$ non-overlapping blocks with a size of $B/2\times B/2$.
		\STATE Estimate the noise variance in each new decomposed block using PCA-based method, with $ \left \{ \left ( m_n,s_n \right )|n\in\left \{ 1,2,\cdots,4N \right \} \right \}$ obtained.
		\STATE Calculating the image tampered probabilities $\mathcal{J}_1$ and $\mathcal{J}_2$ by using (\ref{eq22}) and (\ref{eq25}).
		\STATE Combine two probabilities as $\mathcal{J}_C$ by using (\ref{eq27}) for refinement.
		\STATE Estimate the MRF parameters $\left( \hat{\bm{f}},\,\hat{\bm{\omega}} \right)$ by solving the energy minimization problem (\ref{eq28}) according to iterative alternating strategy.
		\STATE Infer the decision map of forgery localization $\hat{\bm{f}}$ by solving the MAP-MRF optimization problem (\ref{eq20}).
	\end{algorithmic}
	\label{algorithm_2}
\end{algorithm}

\section{Experimental Results on Splicing Location Algorithm}\label{experiments}
In this section, we present experiments conducted on splicing tampered images to test our algorithm. In Sec. \ref{setup}, we introduce the experimental preparation, including image datasets and evaluation criterion. In Sec. \ref{sdn detector}, we carry out some preliminary analysis to assess the performance of SDN in forgery localization. In Sec. \ref{compare}, we evaluate our method in quantitative and qualitative terms and then compare results with some noise-related algorithms. Finally, in Sec. \ref{robustness}, we measure the impact of lossy JPEG compression and scaling attack on images.
\subsection{Experimental Setup}\label{setup}
In order to assess the performance of forgery localization in different scenarios, we used five datasets, listed in Table \ref{table2} together with their main features. The Columbia dataset \cite{hsu2006icme} comprised forged images with splicing, but they were edge-salient and not realistic enough. Similar to the Columbia dataset, the UM-IPPR dataset \cite{umippr} contained forged images with splicing objects by some simple manipulations. The manipulations in the DSO-1 dataset \cite{carvalho2013tifs}, where images were saved in uncompressed PNG format, were carried out with great care, and most of them were realistic. Forgeries of various types were present in the Realistic Tampering Dataset (RTD) proposed by Korus in \cite{korus2017tifs}. The manipulated images, all uncompressed, appear extremely realistic, although only a small number of cameras were involved. Notably, in our experiment, we only chose the forged images with splicing manipulation in the RTD for the target for splicing forgery localization. The USST-SPLC dataset\footnote{https://1drv.ms/u/s!ArHfoaqCoSuIgkKpV9voJuFztoSv?e=HFQpfx} was a self-created dataset containing 56 different tampered images, which were created from authentic ones with different ISO settings using Adobe Photoshop, to approximate the realistic scenario. The tampered region could originate from the same camera with the host image or a different one. All the source pictures were downloaded from the famous camera review website \emph{depeview.com}.
\begin{table}[!t]
	\renewcommand{\arraystretch}{1.3}
	\renewcommand\tabcolsep{1.5pt} 
	\caption{Datasets}
	\label{table2}
	\centering
	\begin{tabular}{ccccc}
		\hline\hline
		Dataset& Ref.& $\#$ Camera& Image Size& format\\ \hline
		Columbia& \cite{hsu2006icme}& 4& $757\times 568-1152\times 768$& TIF\\
		USST-SPLC& self-created& 13& $1920\times 1080-8368\times 5584$& TIF\\
		DSO-1& \cite{carvalho2013tifs}& unknown& $2048\times 1536$& PNG\\
		RTD (Korus)& \cite{korus2017tifs}& 4& $1920\times 1080$& TIF\\
		UM-IPPR& \cite{umippr}& unknown& 1152$\times 768$& JPG\\
		\hline\hline
	\end{tabular}
	\vspace{-0.3cm}
\end{table}

For reference methods, we only considered noise-related methods, listed as NOI1 \cite{mahdian2009ivc}, NOI2 \cite{lyu2014ijcv}, NOI4 \cite{zeng2017mta}, NOI-Multiscale \cite{pun2016jvcir}, NOI-SVD \cite{liu2020neuc}, Splicebuster \cite{cozzolino2015wifs}, and Noiseprint \cite{cozzolino2020tifs}. NOI1 and NOI2 were based on error level analysis and dependent on the noise intensity only. NOI4 and NOI-SVD used the noise estimated in the PCA domain and SVD domain separately to reveal the noise inconsistency in forged images. NOI-Multiscale was a method of adopting the multiscale strategy and distance probability map to cluster the original and tampered region. In Splicebuster, the high-pass noise residual of the image was employed to extract rich features, and then these features were clustered using the EM algorithm to reveal possible anomalies. In comparison, Noiseprint was a CNN-based method to detect the traces of camera artifacts.

To evaluate the performance of SDN and the proposed algorithm in forgery localization, we used \emph{precision}, \emph{recall}, \emph{accuracy}, and \emph{F-score} as assessment criteria, which are defined as follows:
\begin{equation}
\label{eq29}
precision=\frac{TP}{TP+FP},
\end{equation}
\begin{equation}
\label{eq30}
recall=\frac{TP}{TP+FN},
\end{equation}
\begin{equation}
\label{eq31}
accuracy=\frac{TP+TN}{TP+TN+FP+FN},
\end{equation}
\begin{equation}
\label{eq32}
F=\frac{2\times precision\times recall}{precision+recall},
\end{equation}
where \emph{TP}, \emph{TN}, \emph{FP}, and \emph{FN} denote the statistics of the observed true positives, true negatives, false positives, and false negatives, respectively.

\subsection{Preliminary Analysis of SDN detector}\label{sdn detector}
\begin{figure}[!t]
	\centering
	\subfloat{\includegraphics[width=1.2in]{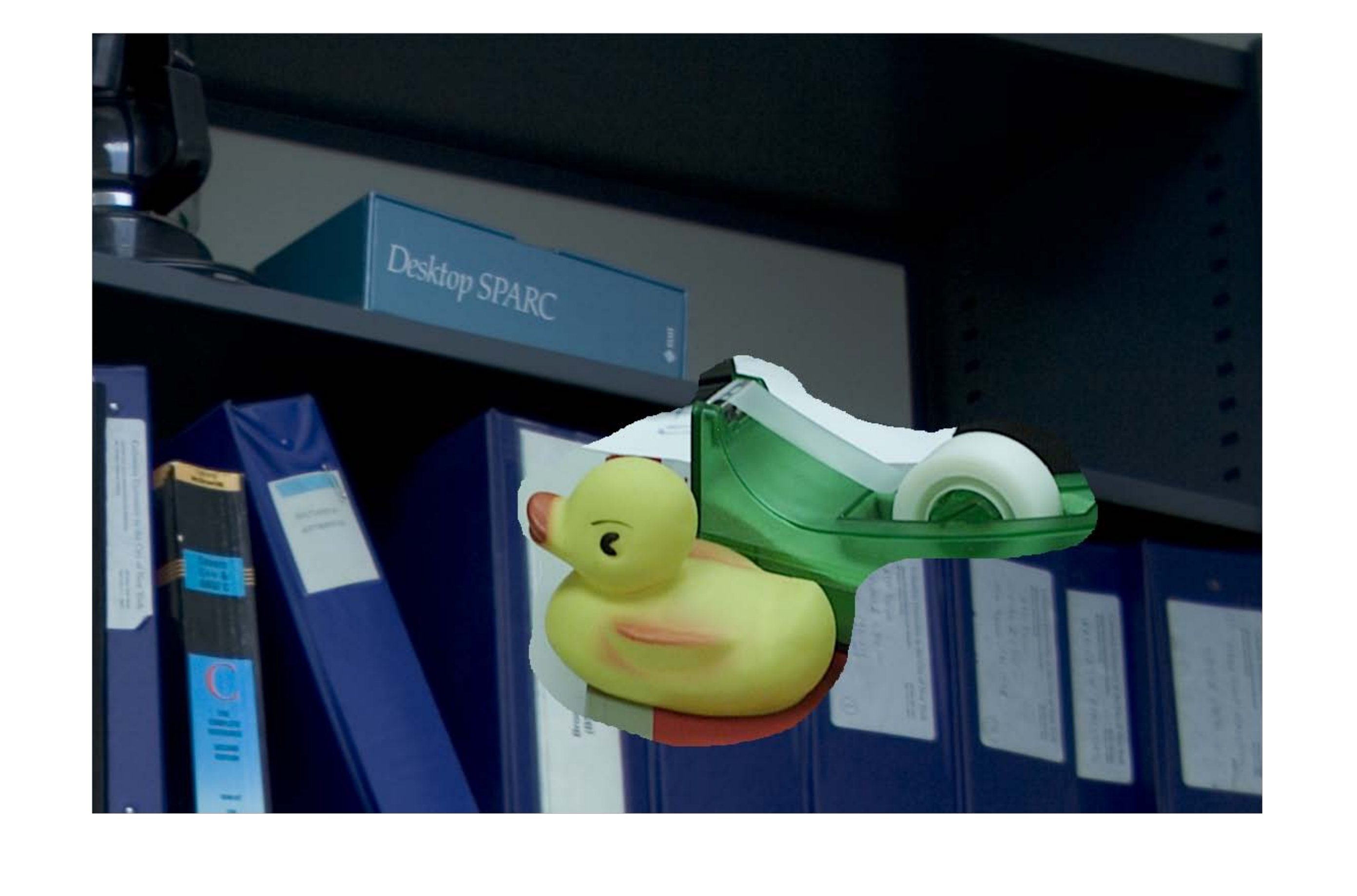}}
	\subfloat{\includegraphics[width=1.2in]{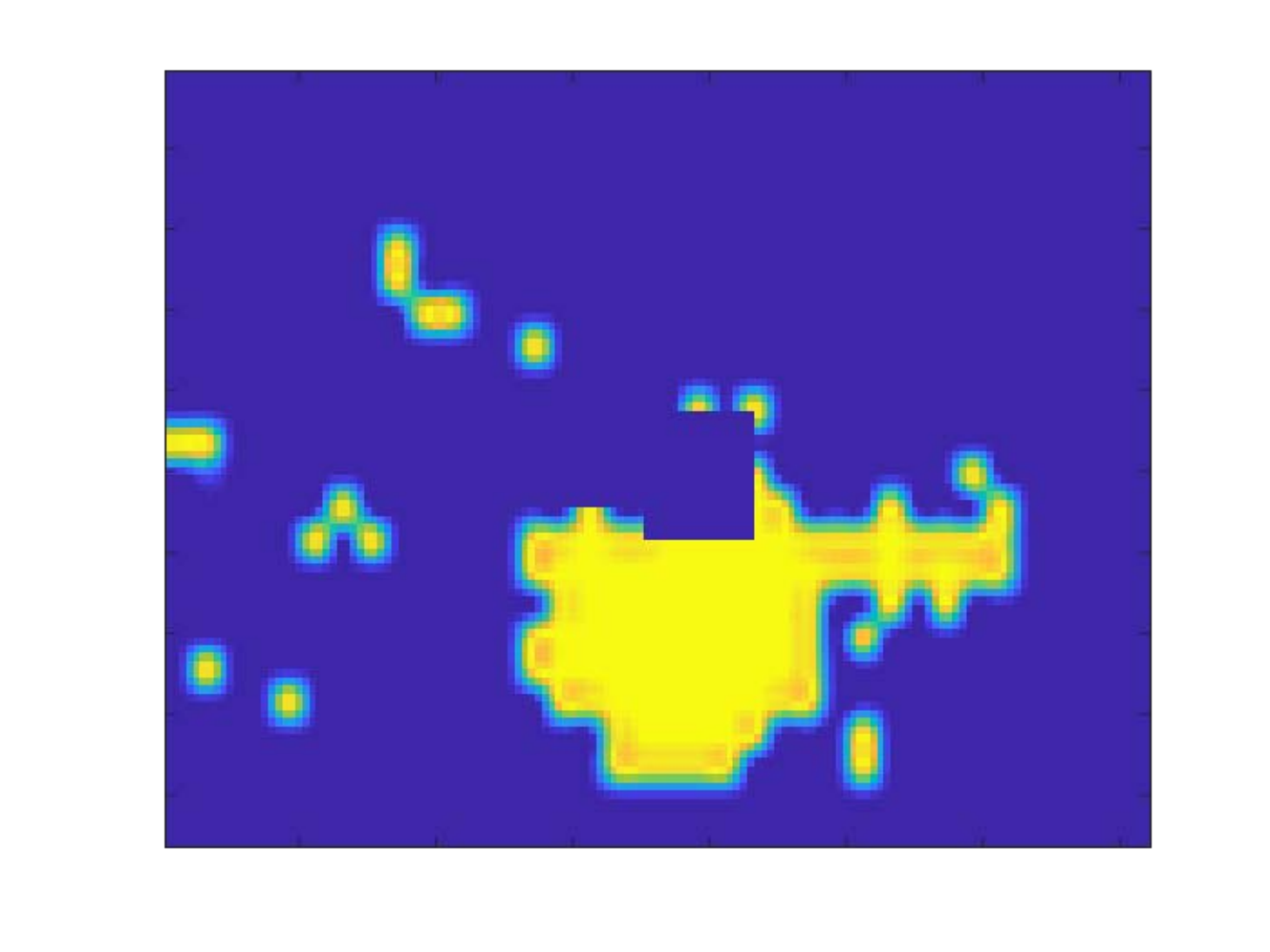}}
	\subfloat{\includegraphics[width=1.2in]{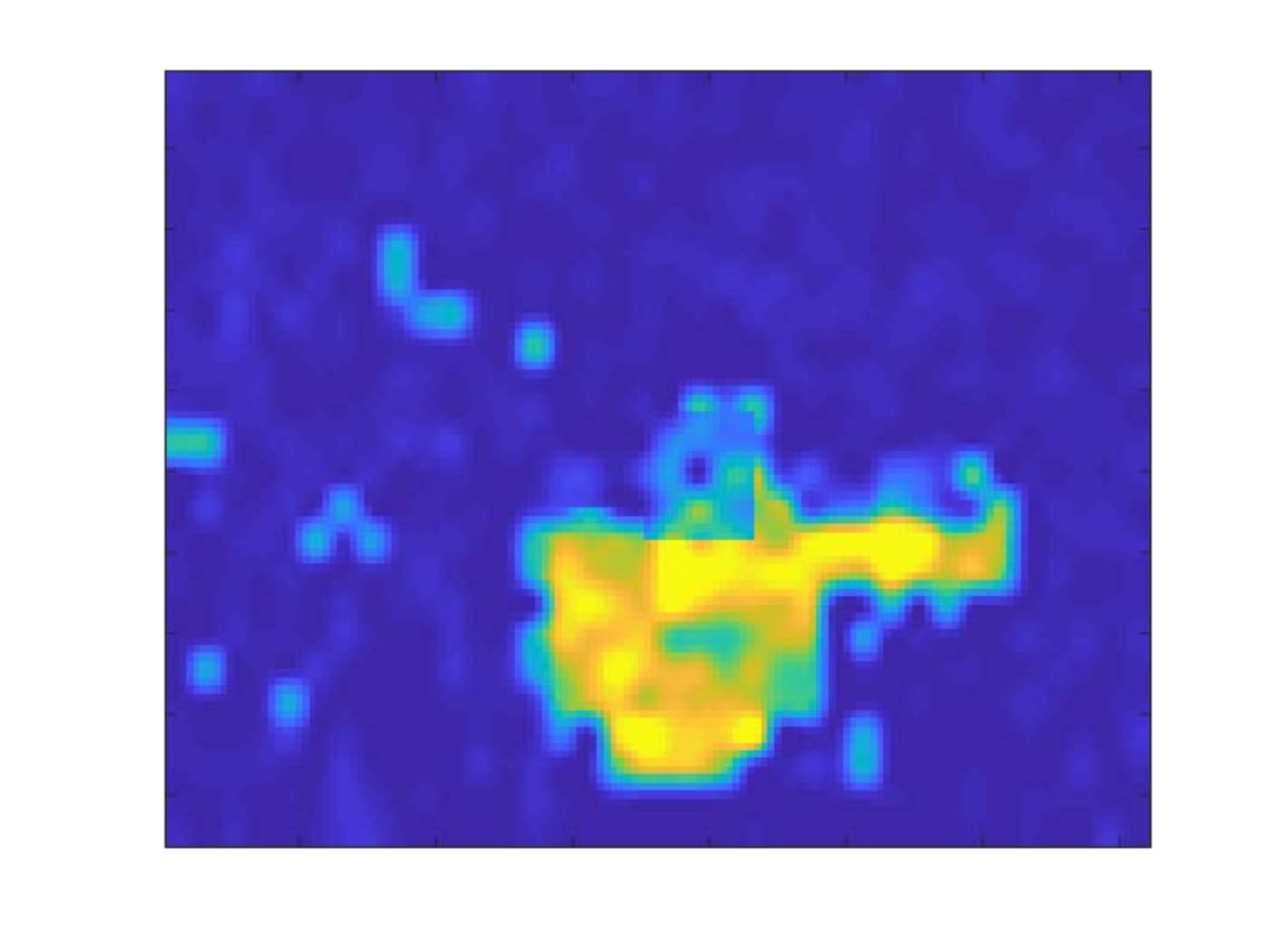}}
	\vspace{-0.3cm}
	\quad
	\centering
	\setcounter{subfigure}{0} 
	\subfloat[]{\includegraphics[width=1.2in]{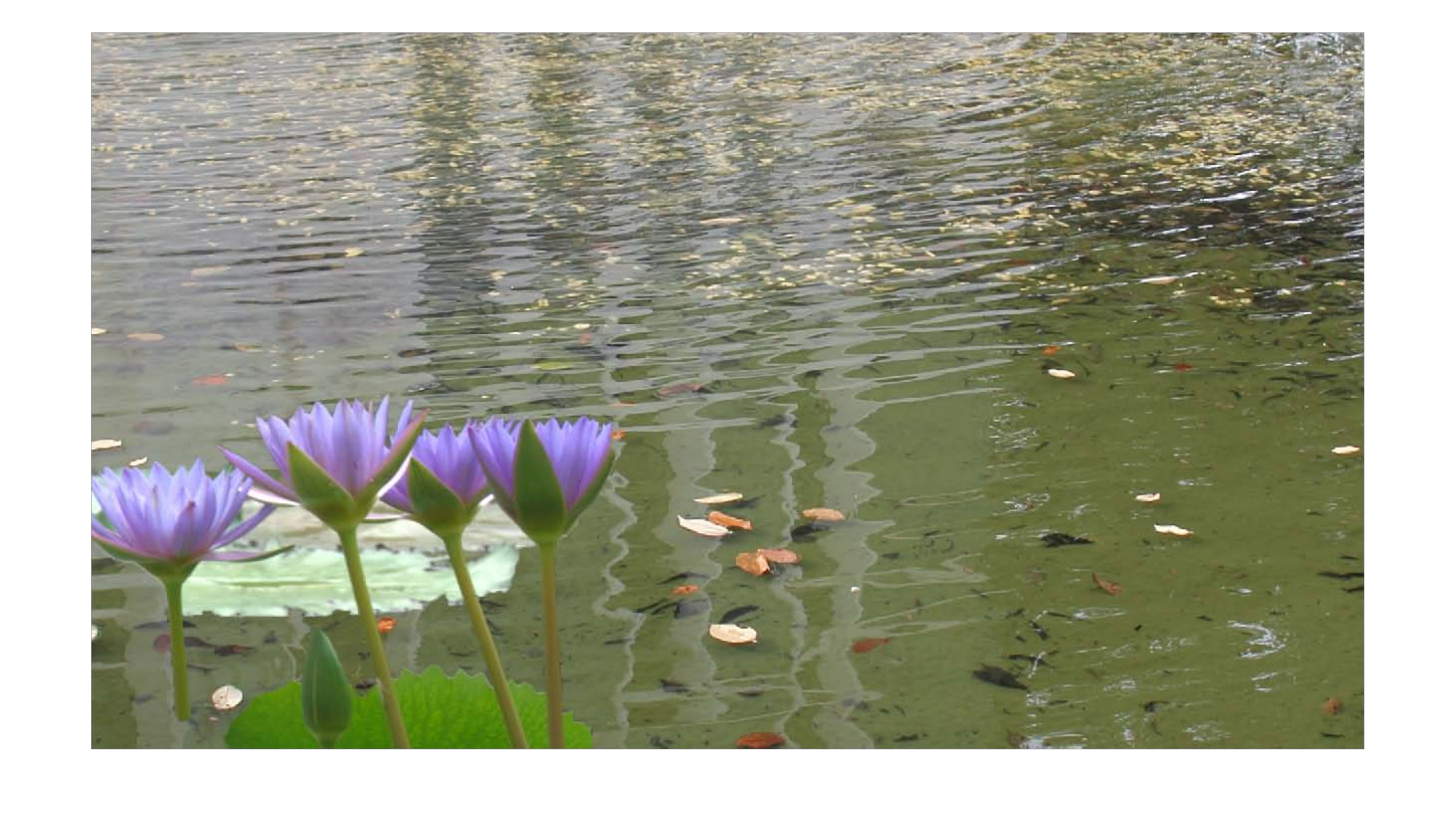}}
	\subfloat[]{\includegraphics[width=1.2in]{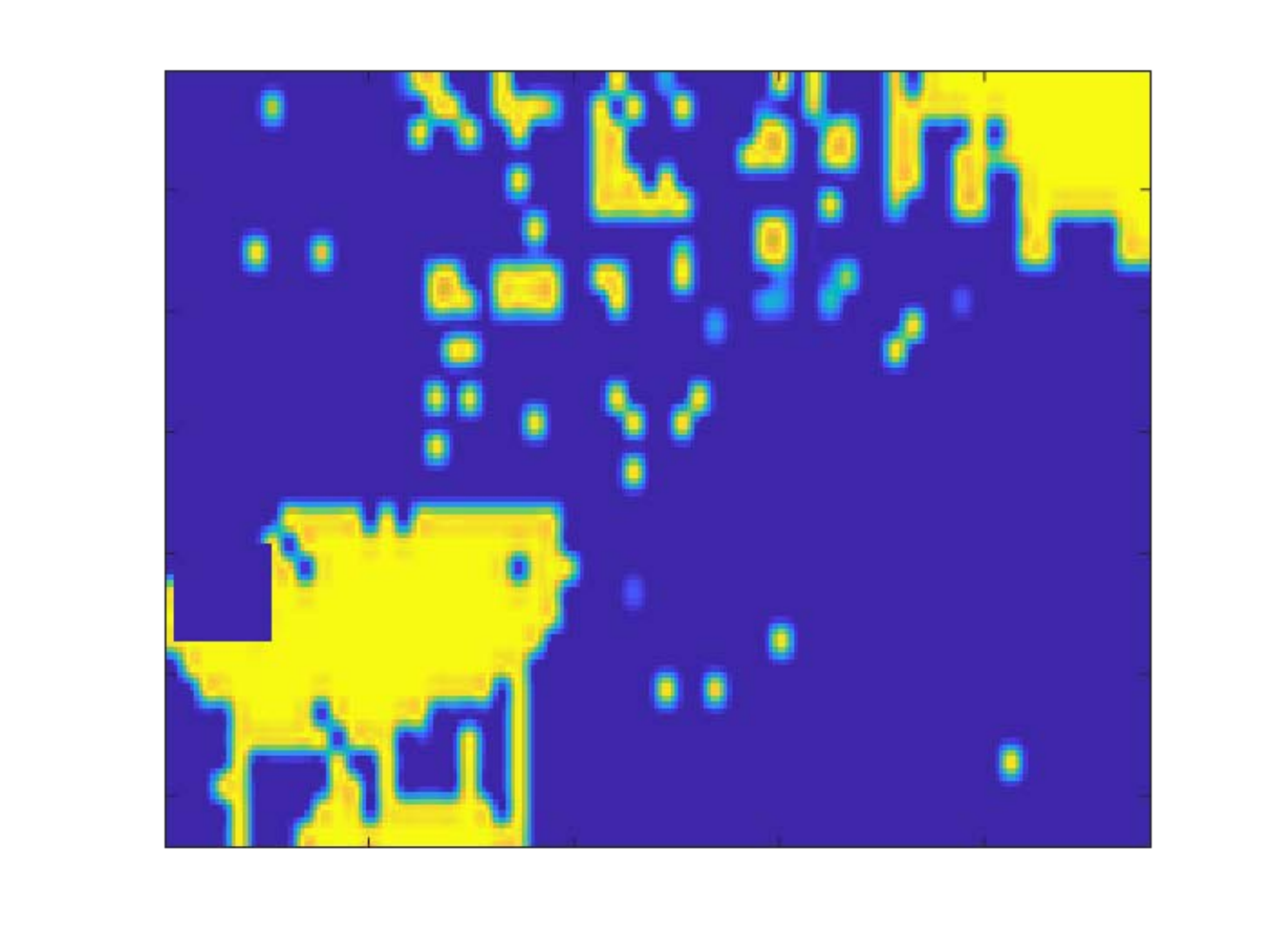}}
	\subfloat[]{\includegraphics[width=1.2in]{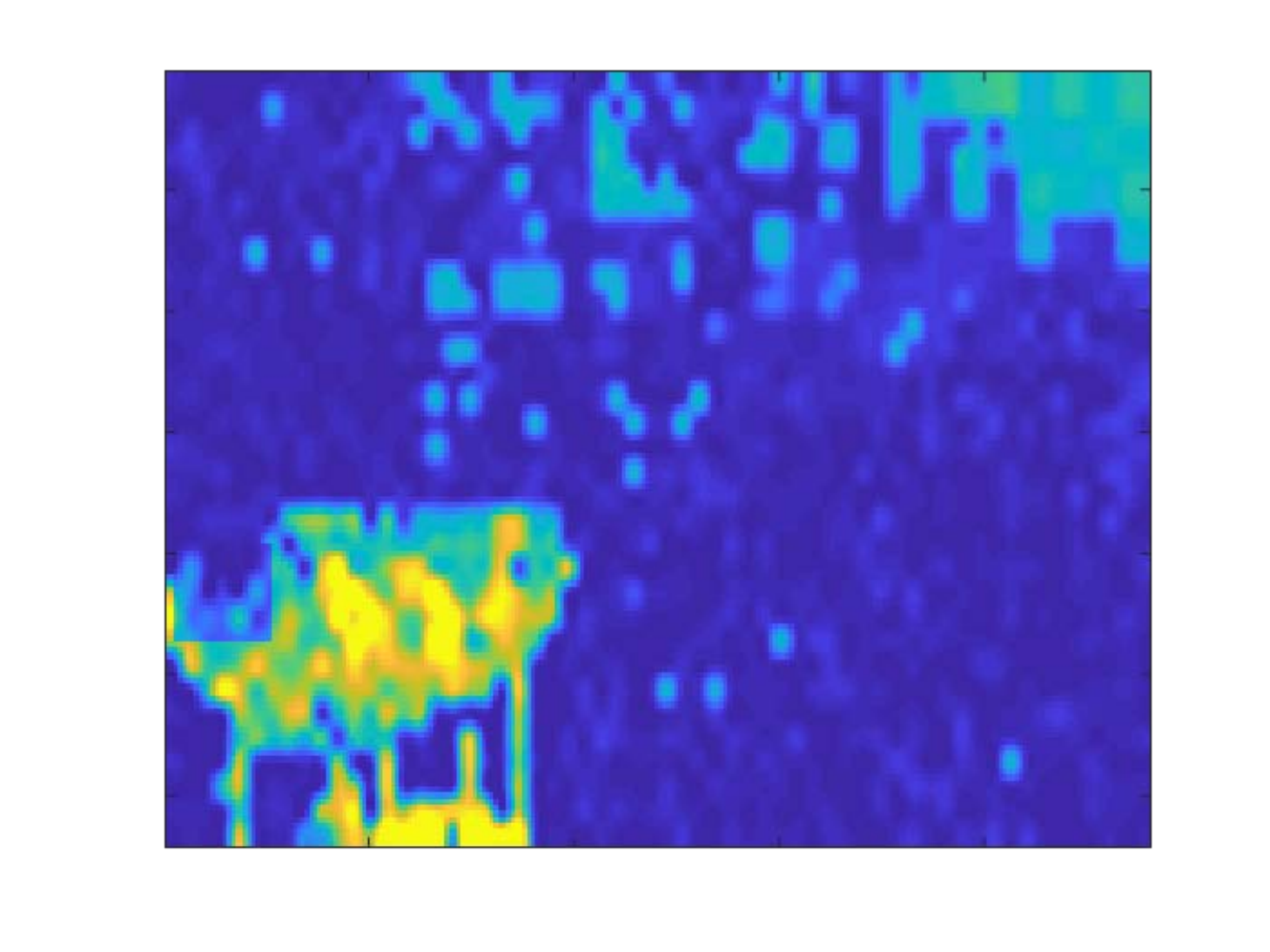}}
	\caption{(a) Forged images; (b) SDN-based tampered maps without refinement; (c) SDN-based tampered maps with refinement.}
	\label{fig7}
	\vspace{-0.3cm}
\end{figure}

Before proceeding to comparative experiments, we carried out a preliminary analysis to assess the performance of SDN in forgery localization. In Sec. \ref{iii-d}, we made a refinement of distance-based probability. Fig. \ref{fig7} illustrates the result of SDN-based tampered maps with and without refinement. On the one hand, the SDN-based tampered maps mistook some of the original region as tampered or omitted some of the tampered area, which would degrade the localization performance of a detector or an analyst’s work. On the other hand, the SDN-based tampered map without refinement provided a relatively definite probability map, subsequently resulting in a hard classification. Conversely, after refinement, the probability map could lead to a soft classification. More precisely, we plotted precision-recall (PR) curves with and without refinement to show the difference. To this end, we selected the Columbia dataset and the USST-SPLC dataset for tests. For each value of recall, we plotted the average precision over all the tested images in the PR curve, as shown in Fig. 8. The refinement-based detector clearly demonstrated a better localization performance, on the whole, than the SDN-based detector without refining.

For completeness, in Fig. \ref{fig9}, we have also provided results in terms of receiver operating characteristic (ROC) curves for the SDN-based probability method without the MRF model. In each curve, for each value of the false positive rate (FPR), the average true positive rate (TPR) over all the tested images is plotted. Note that, considering NOI4, NOI-Multiscale, and NOI-SVD provided binary detection results, but were not suitable for ROC curves. It is easily observed in Fig. \ref{fig9} that SDN detector provided a stable and competitive performance as a classifier, though it was not the best in some datasets. Note that the provided results in Fig. \ref{fig9} were conducted by a simply binary classification without any assistive means and here is only an evaluation of the SDN detector in forgery localization. Fig. \ref{fig9} indicates that the SDN-based detector has potential to identify and locate the tampered region in an image, but may need some additive tools to improve the performance.
\begin{figure}[!t]
	\centering
	\subfloat[Columbia dataset]{\includegraphics[width=1.52in]{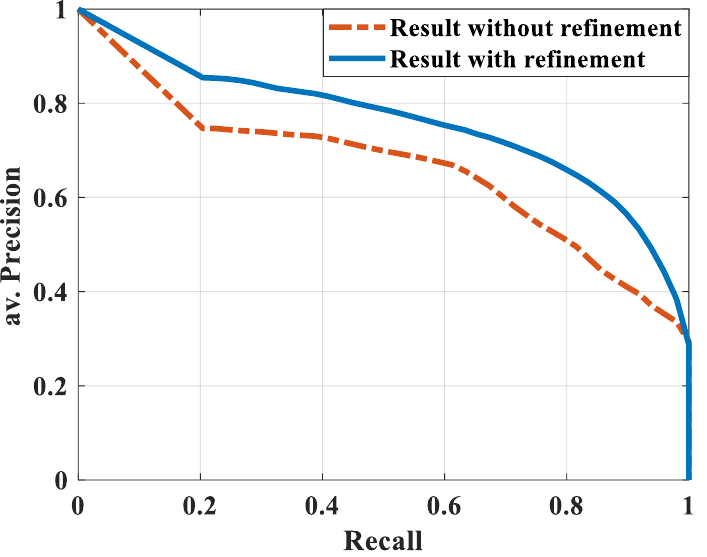}}\quad
	\subfloat[USST-SPLC dataset]{\includegraphics[width=1.5in]{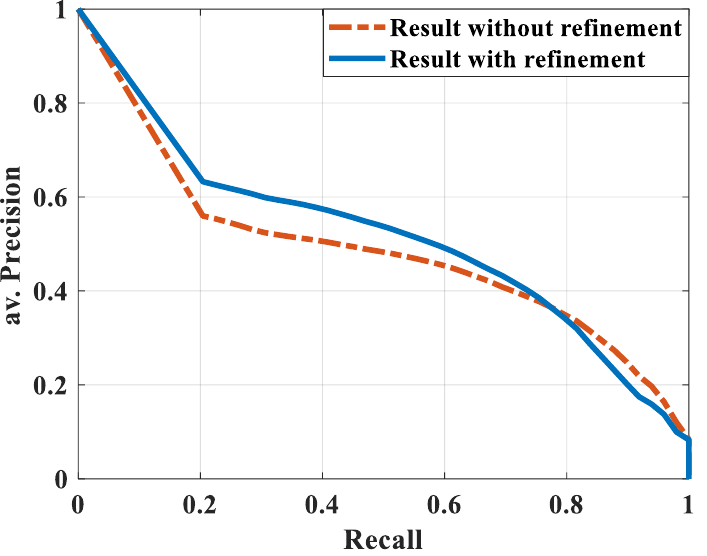}}
	\caption{Precision-Recall curves of two different datasets.}
	\label{fig8}
	\vspace{-0.5cm}
\end{figure}
\begin{figure*}[!t]
	\centering
	\subfloat[Columbia dataset]{\includegraphics[width=1.3in]{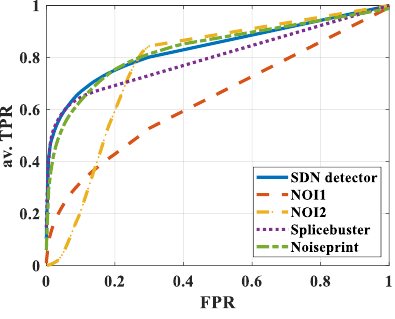}}\quad
	\subfloat[DSO-1]{\includegraphics[width=1.3in]{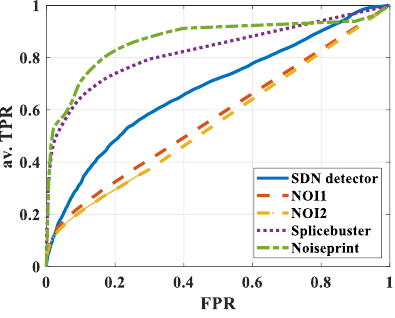}}\quad
	\subfloat[RTD (Korus)]{\includegraphics[width=1.3in]{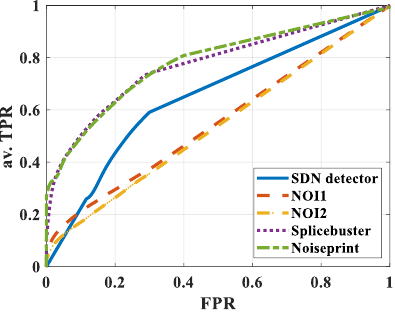}}\quad
	\subfloat[USST-SPLC]{\includegraphics[width=1.3in]{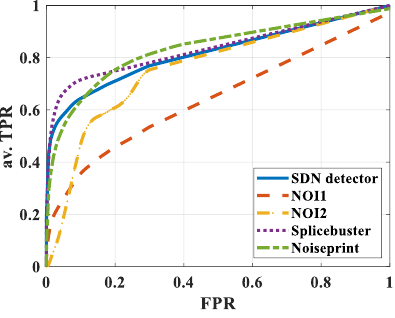}}
	\\
	\centering
	\subfloat[UM-IPPR-1]{\includegraphics[width=1.3in]{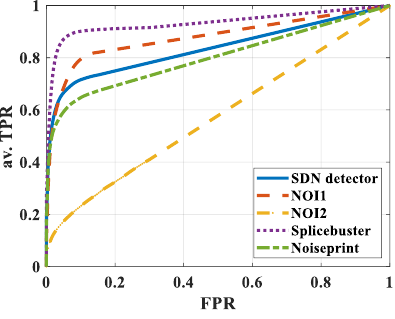}}\quad
	\subfloat[UM-IPPR-3]{\includegraphics[width=1.3in]{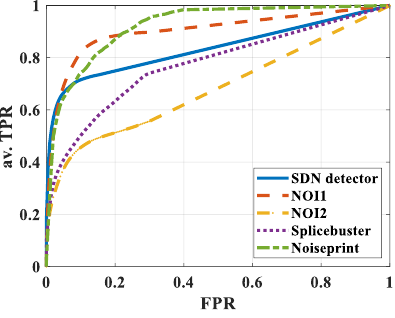}}\quad
	\subfloat[UM-IPPR-5]{\includegraphics[width=1.3in]{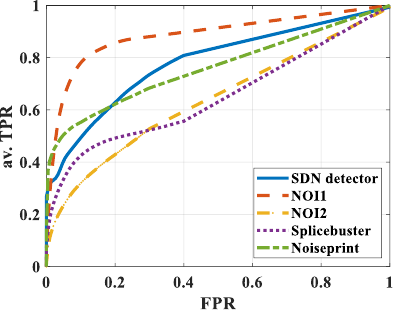}}
	\caption{ROC curves of SDN detector for all datasets.}
	\label{fig9}
\end{figure*}
\vspace{-0.2cm}
\subsection{Comparative Experiments in Forgery Localization}\label{compare}
In this subsection, we will evaluate the localization performance of a refinement-based SDN detector with the MRF labeling model, presenting the results of 8 methods for all 5 datasets presented in Tables \ref{table3}-\ref{table5}, for \emph{precision}, \emph{recall}, and \emph{F-score}, respectively. For a synoptic view of the performance, we have complemented each performance value with the corresponding rank on the dataset, in parentheses, using red for the three best methods and blue for the others. The last column shows the average ranking over all datasets. Considering that the UM-IPPR dataset comprises one spliced object or multiple objects in the forged images, here we chose 1, 3, and 5 objects in the experiment to analyze the influence of multiple objects. Note that in contrast to NOI4, NOI-Multiscale, and NOI-SVD, the other compared methods provided a heatmap rather than a binary localization map for assessment. Hence, we used K-means clustering algorithm to localize the forged region in these heatmaps for the experiments and unified evaluation.
\begin{table*}[!t]
	\renewcommand{\arraystretch}{1.3}
	\renewcommand\tabcolsep{2.0pt} 
	\caption{Experimental Results: \emph{Precision}}
	\label{table3}
	\centering
	\begin{tabular}{ccccccccc}
		\hline\hline
		Dataset& Columbia& USST-SPLC& DSO-1& RTD (Korus)& UM-IPPR-1& UM-IPPR-3& UM-IPPR-5& Av. Rank\\ \hline
		Proposed& 0.7853\textcolor{red}{(1)}& 0.5597\textcolor{red}{(3)}& 0.3704\textcolor{red}{(3)}& 0.2302\textcolor{red}{(3)}& 0.6496\textcolor{red}{(3)}& 0.8059\textcolor{red}{(3)}& 0.8457\textcolor{red}{(3)}& 2.7\textcolor{red}{(1)}\\
		
		NOI1& 0.5263\textcolor{blue}{(5)}& 0.2034\textcolor{blue}{(7)}& 0.1307\textcolor{blue}{(7)}& 0.1770\textcolor{blue}{(8)}& 0.6191\textcolor{blue}{(4)}& 0.9627\textcolor{red}{(1)}& 0.9702\textcolor{red}{(1)}& 4.7\textcolor{blue}{(6)} \\
		
		NOI2& 0.4451\textcolor{blue}{(7)}& 0.2158\textcolor{blue}{(6)}& 0.1873\textcolor{blue}{(6)}& 0.1340\textcolor{blue}{(7)}& 0.2607\textcolor{blue}{(8)}& 0.4038\textcolor{blue}{(8)}& 0.4952\textcolor{blue}{(6)}& 6.9\textcolor{blue}{(8)} \\
		
		NOI4& 0.6732\textcolor{blue}{(4)}& 0.3022\textcolor{blue}{(4)}& 0.1243\textcolor{blue}{(8)}& 0.1010\textcolor{blue}{(6)}& 0.5977\textcolor{blue}{(5)}& 0.9456\textcolor{red}{(2)}& 0.9612\textcolor{red}{(2)}& 4.4\textcolor{blue}{(5)} \\
		
		NOI-Multiscale& 0.4293\textcolor{blue}{(8)}& 0.1396\textcolor{blue}{(8)}& 0.2147\textcolor{blue}{(4)}& 0.1169\textcolor{blue}{(4)}& 0.3107\textcolor{blue}{(7)}& 0.4805\textcolor{blue}{(6)}& 0.5360\textcolor{blue}{(5)}& 6.0\textcolor{blue}{(7)} \\
		
		NOI-SVD& 0.4832\textcolor{blue}{(6)}& 0.2932\textcolor{blue}{(5)}& 0.2069\textcolor{blue}{(5)}& 0.1062\textcolor{blue}{(5)}& 0.7529\textcolor{red}{(1)}& 0.7989\textcolor{blue}{(4)}& 0.7773\textcolor{blue}{(4)}& 4.3\textcolor{blue}{(4)} \\
		
		Splicebuster& 0.6934\textcolor{red}{(3)}& 0.5885\textcolor{red}{(1)}& 0.6197\textcolor{red}{(2)}& 0.3639\textcolor{red}{(1)}& 0.7154\textcolor{red}{(2)}& 	0.4596\textcolor{blue}{(7)}& 0.2003\textcolor{blue}{(8)}& 3.4\textcolor{red}{(2)} \\
		
		Noiseprint& 0.6989\textcolor{red}{(2)}& 0.5719\textcolor{red}{(2)}& 0.7534\textcolor{red}{(1)}& 0.3418\textcolor{red}{(2)}& 0.5769\textcolor{blue}{(6)}& 0.5181\textcolor{blue}{(5)}& 0.3961\textcolor{blue}{(7)}& 3.6\textcolor{red}{(3)} \\
		
		\hline\hline
	\end{tabular}
	\vspace{-0.3cm}
\end{table*}
\begin{table*}[!t]
	\renewcommand{\arraystretch}{1.3}
	\renewcommand\tabcolsep{2.0pt} 
	\caption{Experimental Results: \emph{Recall}}
	\label{table4}
	\centering
	\begin{tabular}{ccccccccc}
		\hline\hline
		Dataset& Columbia& USST-SPLC& DSO-1& RTD (Korus)& UM-IPPR-1& UM-IPPR-3& UM-IPPR-5& Av. Rank\\ \hline
		
		Proposed& 0.7598\textcolor{red}{(1)}& 0.8180\textcolor{red}{(1)}& 0.6714\textcolor{red}{(2)}& 0.4471\textcolor{red}{(2)}& 0.6941\textcolor{blue}{(5)}& 0.6982\textcolor{blue}{(5)}& 0.7493\textcolor{blue}{(4)}& 2.9\textcolor{red}{(1)}\\
		
		NOI1& 0.3963\textcolor{blue}{(4)}& 0.4671\textcolor{blue}{(7)}& 0.2039\textcolor{blue}{(7)}& 0.3710\textcolor{blue}{(6)}& 0.7807\textcolor{red}{(3)}& 0.8895\textcolor{red}{(1)}& 0.8680\textcolor{red}{(1)}& 4.1\textcolor{blue}{(4)}\\
		
		NOI2& 0.2750\textcolor{blue}{(7)}& 0.4819\textcolor{blue}{(6)}& 0.1064\textcolor{blue}{(8)}& 0.1333\textcolor{blue}{(8)}& 0.3104\textcolor{blue}{(7)}& 0.3002\textcolor{blue}{(8)}& 0.2808\textcolor{blue}{(6)}& 7.1\textcolor{blue}{(8)}\\
		
		NOI4& 0.5778\textcolor{red}{(3)}& 0.6351\textcolor{red}{(3)}& 0.2060\textcolor{blue}{(6)}& 0.3659\textcolor{blue}{(7)}& 0.6314\textcolor{blue}{(6)}& 0.7057\textcolor{blue}{(4)}& 0.6701\textcolor{blue}{(5)}& 4.9\textcolor{blue}{(6)}\\
		
		NOI-Multiscale& 0.3831\textcolor{blue}{(5)}& 0.3083\textcolor{blue}{(8)}& 0.2832\textcolor{blue}{(4)}& 0.4346\textcolor{blue}{(4)}& 0.8851\textcolor{red}{(1)}& 0.8489\textcolor{red}{(3)}& 0.8158\textcolor{red}{(2)}& 3.9\textcolor{red}{(2)}\\
		
		NOI-SVD& 0.1086\textcolor{blue}{(8)}& 0.4867\textcolor{blue}{(5)}& 0.2433\textcolor{blue}{(5)}& 0.4413\textcolor{red}{(3)}& 0.8809\textcolor{red}{(2)}& 0.8490\textcolor{red}{(2)}& 0.8020\textcolor{red}{(3)}& 4.0\textcolor{red}{(3)}\\
		
		Splicebuster& 0.3490\textcolor{blue}{(6)}& 0.6132\textcolor{blue}{(4)}& 0.4526\textcolor{red}{(3)}& 0.4092\textcolor{blue}{(5)}& 0.7796\textcolor{blue}{(4)}& 0.3263\textcolor{blue}{(6)}& 0.1123\textcolor{blue}{(7)}& 5.0\textcolor{blue}{(7)}\\
		
		Noiseprint& 0.6536\textcolor{red}{(2)}& 0.6825\textcolor{red}{(2)}& 0.7078\textcolor{red}{(1)}& 0.4472\textcolor{red}{(1)}& 0.2754\textcolor{blue}{(8)}& 0.3026\textcolor{blue}{(7)}& 0.1108\textcolor{blue}{(8)}& 4.1\textcolor{blue}{(4)}\\
		
		\hline\hline
	\end{tabular}
	\vspace{-0.3cm}
\end{table*}

Table \ref{table3} shows clearly that the proposed method obtained the best result of precision in an average ranking. From Table \ref{table3}, despite the average ranking, the proposed method achieved the best performance for the Columbia dataset, with \emph{precision} = 0.7853, nearly $9\%$ higher than the second best (Noiseprint) and much better than the other methods. In the 3 subsets of the UM-IPPR dataset, the values of \emph{precision} can be seen to rise along with the increasing of number of spliced objects. This outcome was mainly due to the fact that more spliced objects were beneficial for recovering NLF of an alien region. Meanwhile, it is worth noting that the proposed method did not always provide the best performance in terms of \emph{precision}. For example, in the USST-SPLC, DSO-1, and RTD datasets, the proposed method ranked third, after Splicebuster and Noiseprint. In fact, the USST-SPLC, DSO-1, and RTD datasets were more complicated and realistic than the Columbia dataset. Accordingly, NLF inconsistence may be small, which affects the performance of the SDN-based detector. Nevertheless, the proposed method performed much better than the other noise-related methods, which yielded at least $10\%$ lower results than the proposed method. Another specific case is that Splicebuster and Noiseprint showed a dramatic impairment in the UM-IPPR datasets; for example, Splicebuster ranked only seventh, and Noiseprint ranked fifth in the UM-IPPR-3 dataset. Moreover, their performances degraded with an increasing number of spliced objects. One reason might be that the noise in the alien region of the UM-IPPR dataset was added artificially and obeyed Gaussian distribution only. The performance of Splicebuster and Noiseprint here suggests that they fitted the real tampered scene well but performances were limited to scenes where noise was added manually. However, the proposed method consistently provided a satisfactory performance across all the datasets.

Tables \ref{table4} and \ref{table5} report the experimental results for the \emph{recall} and \emph{F-score} metrics. Like Table \ref{table3}, the proposed method averagely ranked first in both the \emph{recall} and \emph{F-score} metrics. In Table \ref{table4}, under the criterion of \emph{recall}, we achieved a better rank across the datasets for real noise, with the first ranking in the USST-SPLC dataset and the second ranking in the DSO-1 and RTD datasets. Notably, Splicebuster, which performed well in terms of precision, showed a dramatic impairment on the whole in the \emph{recall}, meaning it tended to omit some forgery in localization. Table \ref{table5} presents the results of the \emph{F-score} metric, which is a summarized quality of \emph{precision} and \emph{recall}. It basically reflects the general results of \emph{precision} and \emph{recall}, shown in the Tables \ref{table3} and \ref{table4}, with a small change in relative ranking.
\begin{table*}[!t]
	\renewcommand{\arraystretch}{1.3}
	\renewcommand\tabcolsep{2.0pt} 
	\caption{Experimental Results: \emph{F-score}}
	\label{table5}
	\centering
	\begin{tabular}{ccccccccc}
		\hline\hline
		Dataset& Columbia& USST-SPLC& DSO-1& RTD (Korus)& UM-IPPR-1& UM-IPPR-3& UM-IPPR-5& Av. Rank\\ \hline
		
		Proposed& 0.7723\textcolor{red}{(1)}& 0.6646\textcolor{red}{(1)}& 0.4774\textcolor{red}{(3)}& 0.3039\textcolor{red}{(3)}& 0.6711\textcolor{blue}{(4)}& 0.7482\textcolor{blue}{(4)}& 0.7946\textcolor{red}{(2)}& 2.6\textcolor{red}{(1)}\\
		
		NOI1& 0.4521\textcolor{blue}{(5)}& 0.2834\textcolor{blue}{(7)}& 0.1593\textcolor{blue}{(6)}& 0.2397\textcolor{blue}{(4)}& 0.6906\textcolor{red}{(3)}& 0.9247\textcolor{red}{(1)}& 0.9163\textcolor{red}{(1)}& 3.9\textcolor{red}{(3)}\\
		
		NOI2& 0.3400\textcolor{blue}{(7)}& 0.2981\textcolor{blue}{(6)}& 0.1357\textcolor{blue}{(8)}& 0.1336\textcolor{blue}{(8)}& 0.2834\textcolor{blue}{(8)}& 0.3444\textcolor{blue}{(8)}& 0.3584\textcolor{blue}{(6)}& 7.3\textcolor{blue}{(8)}\\
		
		NOI4& 0.6219\textcolor{red}{(3)}& 0.4095\textcolor{blue}{(4)}& 0.1550\textcolor{blue}{(7)}& 0.1583\textcolor{blue}{(7)}& 0.6141\textcolor{blue}{(5)}& 0.8082\textcolor{red}{(3)}& 0.7896\textcolor{red}{(3)}& 4.6\textcolor{blue}{(6)} \\
		
		NOI-Multiscale& 0.4049\textcolor{blue}{(6)}& 0.1922\textcolor{blue}{(8)}& 0.2442\textcolor{blue}{(4)}& 0.1842\textcolor{blue}{(5)}& 0.4599\textcolor{blue}{(6)}& 0.6137\textcolor{blue}{(5)}& 0.6469\textcolor{blue}{(5)}& 5.6\textcolor{blue}{(7)} \\
		
		NOI-SVD& 0.1773\textcolor{blue}{(8)}& 0.3659\textcolor{blue}{(5)}& 0.2177\textcolor{blue}{(5)}& 0.1712\textcolor{blue}{(6)}& 0.8119\textcolor{red}{(1)}& 0.8232\textcolor{red}{(2)}& 0.7895\textcolor{blue}{(4)}& 4.4\textcolor{blue}{(5)} \\
		
		Splicebuster& 0.4643\textcolor{blue}{(4)}& 0.6006\textcolor{red}{(3)}& 0.5231\textcolor{red}{(2)}& 0.3852\textcolor{red}{(2)}& 0.7461\textcolor{red}{(2)}& 0.3816\textcolor{blue}{(7)}& 0.1439\textcolor{blue}{(8)}& 4.0\textcolor{blue}{(4)} \\
		
		Noiseprint& 0.6755\textcolor{red}{(2)}& 0.6223\textcolor{red}{(2)}& 0.7299\textcolor{red}{(1)}& 0.3875\textcolor{red}{(1)}& 0.3728\textcolor{blue}{(7)}& 0.3821\textcolor{blue}{(6)}& 0.1732\textcolor{blue}{(7)}& 3.7\textcolor{red}{(2)}\\
		
		\hline\hline
	\end{tabular}
	\vspace{-0.2cm}
\end{table*}

\begin{figure*}[!t] 
	\centering
	\includegraphics[width=6.5in]{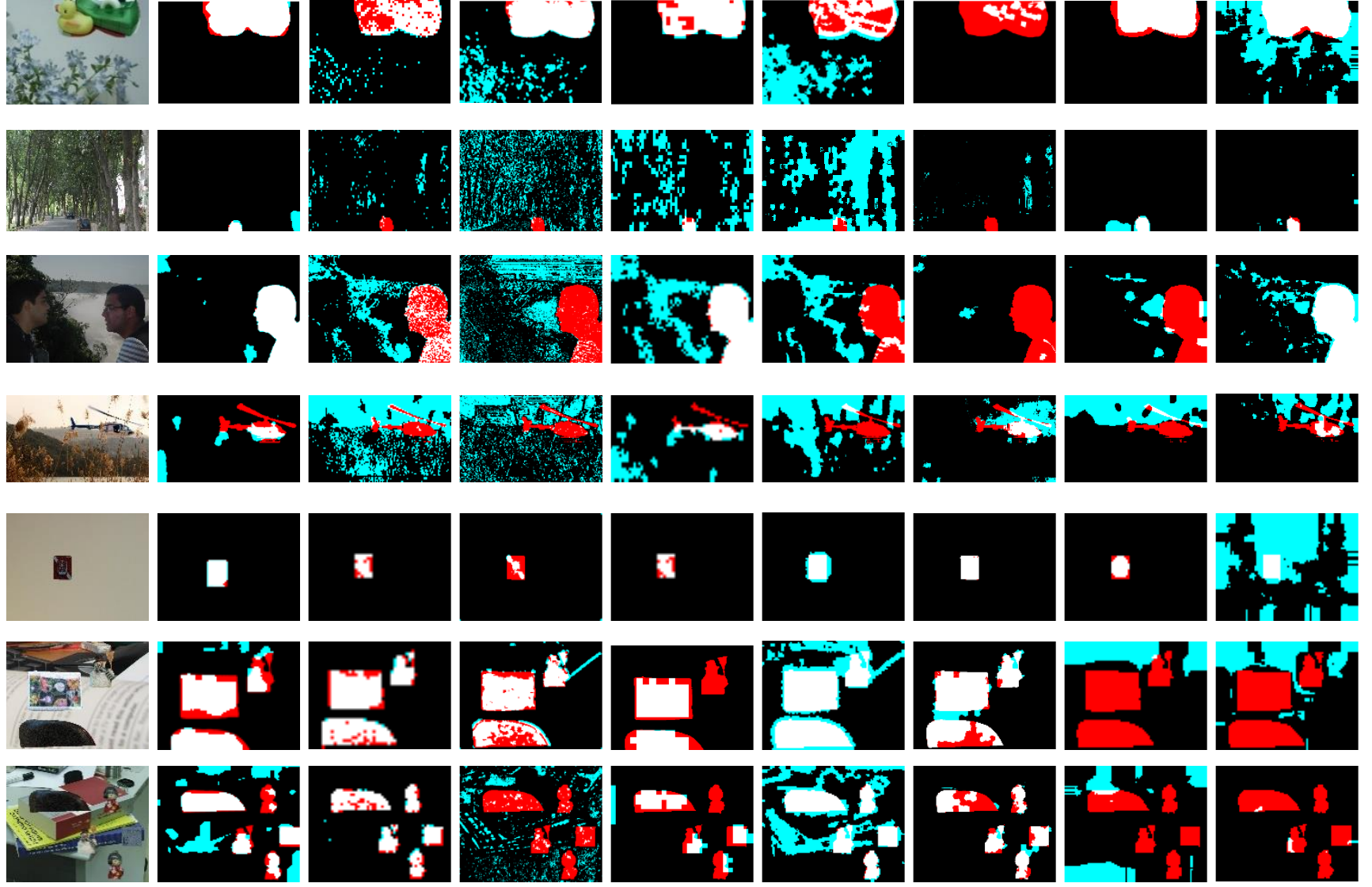}
	\caption{Examples from all the test datasets. From left to right: forged image, color-coded decision maps from the 8 tested methods: proposed, NOI1, NOI2, NOI4, NOI-Multiscale, NOI-SVD, Splicebuster, and Noiseprint.}
	\label{fig10}
	\vspace{-0.5cm}
\end{figure*}

A better understanding of the actual quality of the results can be obtained by visually inspecting the examples in Fig. \ref{fig10}. In this figure, color-coded decision maps display four colors: white, cyan, red, and black, where white denotes the detected tampered region (\emph{TP}), cyan indicates the detected authentic region (\emph{FP}), red designates the undetected tampered region (\emph{FN}), and black signifies the undetected authentic region (\emph{TN}). Notably, Fig. \ref{fig10} demonstrates that the proposed refined SDN-based method with MRF model achieved a better performance of localization though, it still yielded some false positives. It is also worth noting that Splicebuster and Noiseprint provided comparable localization results in the datasets of real scenes but offered relatively poor performance in the UM-IPPR dataset, where noise was manually added in the spliced region. This result also validates the data in Tables \ref{table3}-\ref{table5}.

\subsection{Robustness Evaluation against Common Post-processing Operations}\label{robustness}
\begin{figure}[!t]
	\centering
	\subfloat[Performance under JPEG compression]{\includegraphics[width=1.25in]{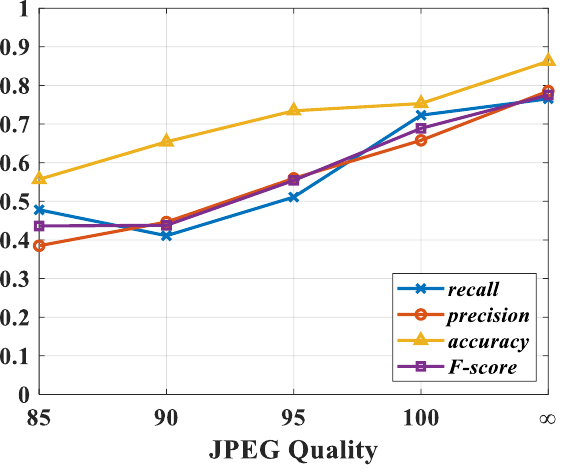}}\quad\quad
	\subfloat[Performance under scaling operation]{\includegraphics[width=1.25in]{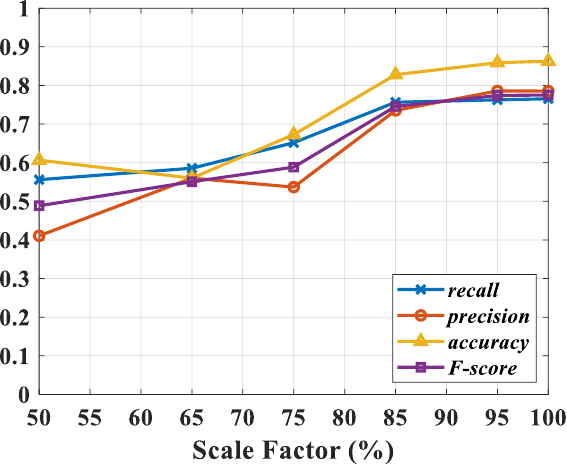}}
	\caption{Impact of lossy JPEG compression and scaling operation on splicing tampered localization performance. Marker at $\infty$ in (a) corresponds to uncompressed TIFF images, and 100\% in (b) means the original images without scaling.}
	\label{fig11}
	\vspace{-0.5cm}
\end{figure}
Considering that the spliced images might have undergone some post-processing in network transmission, we studied the impact of these attacks on our proposed method, using comparisons with other noise-based algorithms. Due to the commonplace of JPEG compression and scaling operation during network transmission, lossy JPEG compression and scaling operation were applied to the Columbia dataset to test the robustness of the proposed splicing detection method. The JPEG compression level was measured by the quality factor, a positive integer in the range of $[1, 100]$. A larger quality factor means higher image quality (i.e., less compression), and vice versa. The scaling operation was measured by a scale factor, which was within the interval of $(0,1]$; a scale factor of 1 meant the original image without any scaling. In this experiment, we created 4 new versions of each image from the Columbia dataset with JPEG quality factors $85$, $90$, $95$, and $100$, and 5 new versions of each image with scale factors $50\%$, $65\%$, $75\%$, $85\%$, and $95\%$.

First, we evaluated the performance of our method for an attack by JPEG compression or scaling. Fig. \ref{fig11} illustrates the results, reflecting the criteria of \emph{recall}, \emph{precision}, \emph{accuracy}, and \emph{F-score}. In Fig. \ref{fig11}, it can be observed that all the criteria for both types of attack decreased when the factors decreased. However, the decreases in the two post-processing were slightly different. In Fig. \ref{fig11}(a), when the JPEG quality factor decreased, \emph{precision} and \emph{recall} dramatically decreased, resulting in a decrease of the \emph{F-score}. Accuracy was sustained when the quality factor was higher than 95, but descended immediately and dramatically when the quality factor was lower than 95. Conversely, performance under the scaling attack remained robust, to some extent, when the scaling factor was larger than 80, as presented in Fig. \ref{fig11}(b). Moreover, it did not result in a downtrend like that seen for JPEG compression, but presented a gradual and even flat decline. For example, as shown in Fig. \ref{fig11}(b), within the interval of scaling factors $[65\%, 75\%]$, the performance of our algorithm remained consistent. On the whole, the proposed method was able to sustain robustness in the scaling operation when the scaling factor was within a certain range. In contrast, JPEG compression destroyed the noise characteristic of an image, making it likely to degrade the performance of the proposed method.
\begin{figure*}[!t]
	\centering
	\subfloat[\emph{precision}]{\includegraphics[width=3in]{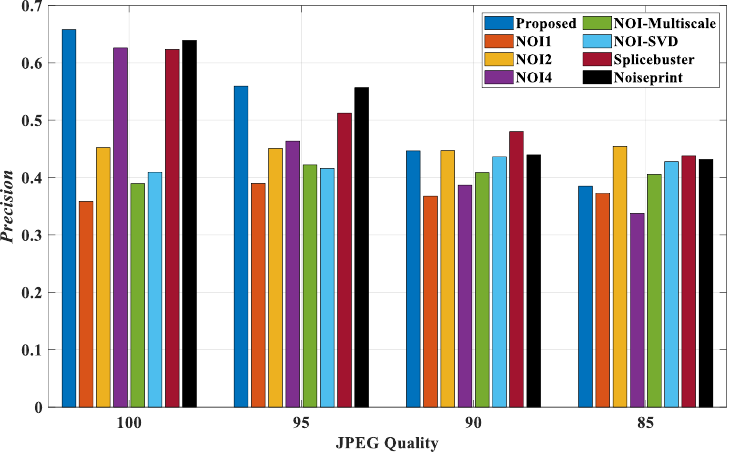}}\quad\quad
	\subfloat[\emph{F-score}]{\includegraphics[width=3in]{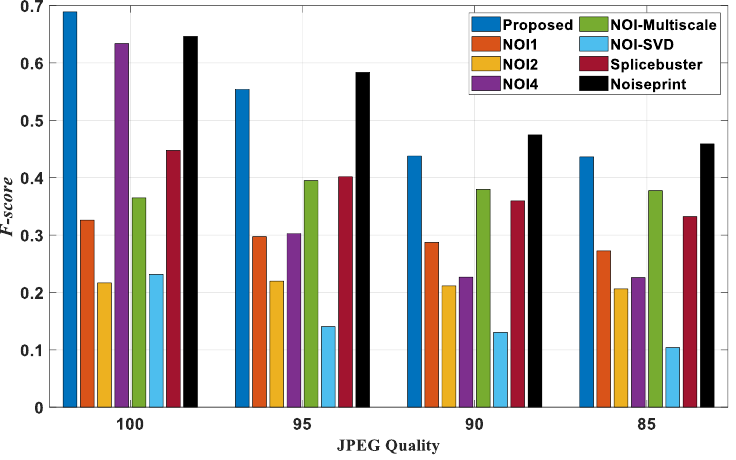}}
	\quad
	\centering
	\subfloat[\emph{recall}]{\includegraphics[width=3in]{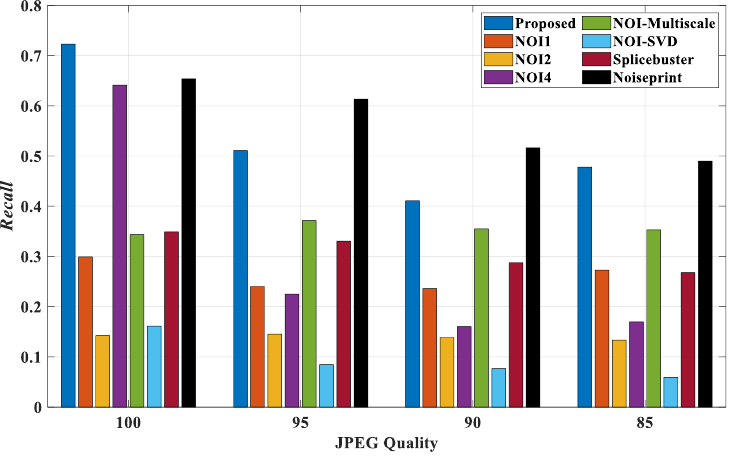}}\quad\quad
	\subfloat[\emph{accuracy}]{\includegraphics[width=3in]{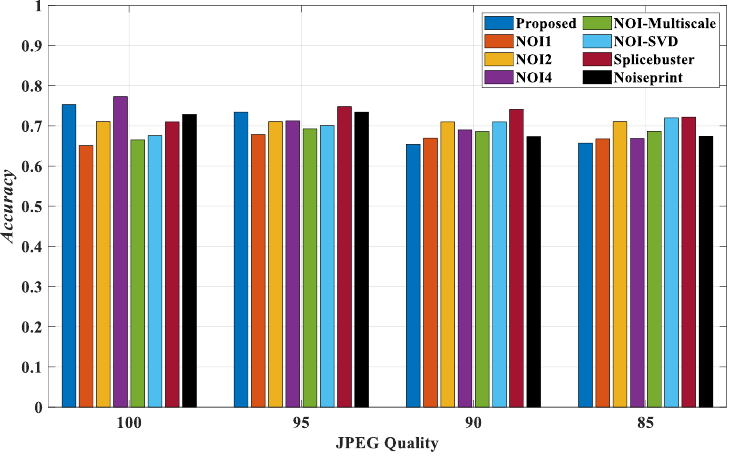}}
	\caption{Robustness comparisons on lossy JPEG compression.}
	\label{fig12}
	\vspace{-0.2cm}
\end{figure*}
\begin{figure*}[!t]
	\centering
	\subfloat[\emph{precision}]{\includegraphics[width=3in]{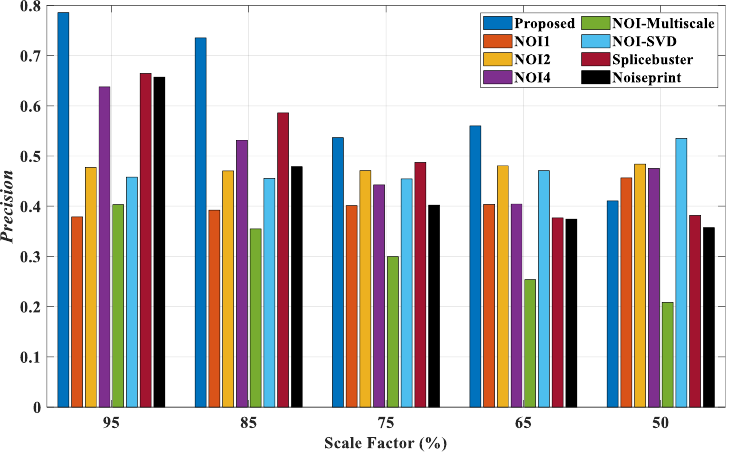}}\quad\quad
	\subfloat[\emph{F-score}]{\includegraphics[width=3in]{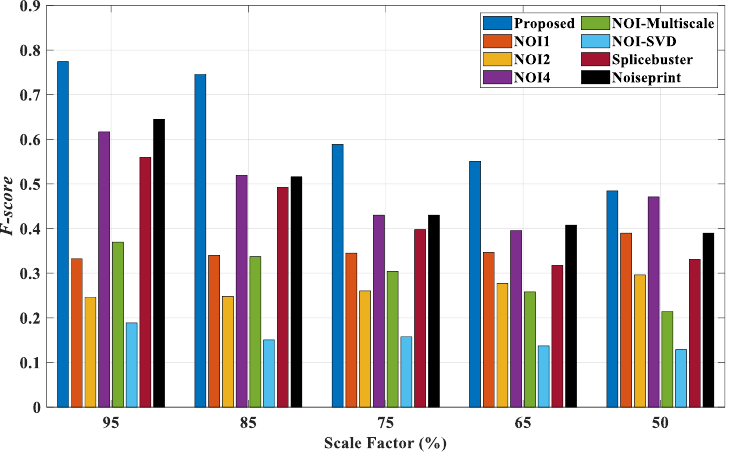}}
	\quad
	\centering
	\subfloat[\emph{recall}]{\includegraphics[width=3in]{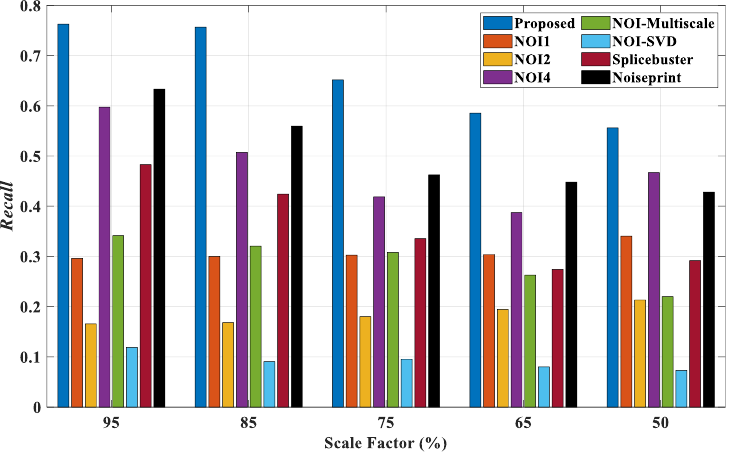}}\quad\quad
	\subfloat[\emph{accuracy}]{\includegraphics[width=3in]{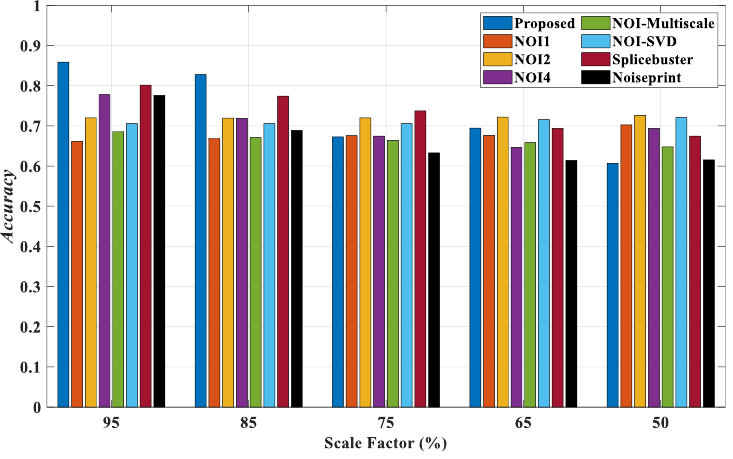}}
	\caption{Robustness comparisons on scaling operation.}
	\label{fig13}
	\vspace{-0.5cm}
\end{figure*}

Next, we compared our algorithm with other noise-inconsistent-based algorithms for localization performance under the attacks of lossy JPEG compression and scaling operation. Comparisons are exhibited in Figs. \ref{fig12} and \ref{fig13}. Figs. \ref{fig12} and \ref{fig13} reveal that the performances of all the methods decreased as the quality factor declined, and the proposed method with different quality factors provided superior or comparative experimental results, especially when quality factors were high. Meanwhile, Fig. 13 clearly demonstrates that our performance on scaling was little changed within the interval above the factor of $75\%$ and between the factors of $65\%$ and $75\%$ but decreased when the factor dropped from $75\%$ to $65\%$. Note that in Figs. \ref{fig12} and \ref{fig13}, accuracies regarding both attacks for all methods are similar and consistent for different parameters. This outcome is mainly due to the fact that the TN (corresponding to a pristine region in a tampered image) of detection was dominant.

\section{Conclusion}\label{conclusion}
This work proposed a novel algorithm for locating a splicing forgery in digital images by applying the likelihood model of NLFs. We first analyzed the statistical model of signal-dependent noise in the empirical evidence and then proposed a likelihood model for NLF. Based on this noise model, we introduced the MAP-MRF framework to detect a splicing forgery in a blind scenario, with conditional probability refinement via a combination strategy. Furthermore, in the MRF framework, we used an iterative alternating approach to estimate MRF parameters, causing our algorithm to be a totally blind detection method without any prior information acquired in advance. Not surprisingly, with consideration of NLF and physical properties in the neighborhood of image space by MRF, the localization performance was improved. Experiments on datasets validated the truth, showing that our method provided superior or comparative experimental results compared with other noise-based methods, both quantitatively and qualitatively.

\appendices
\section{Discussion Concerning Conditional Probability Model}\label{appendix}
Generally, the noise samples on the one side of an NLF curve do not require correction. In contrast, detection based on the conditional probability model may differ slightly from the truth concerning the noise samples between the two curves. Here, we offer a simple discussion.

Consider $\varepsilon _1$ and $\varepsilon _2$ as the distance between a noise sample $(m_n,s_n )$ and two NLFs $\sigma_1$ and $\sigma_2$, where the subscripts 1 and 2 represent NLFs obtained from the so-called authentic and splicing regions, respectively, that is, $\varepsilon_i=\|s_n,\sigma_i \|_2=|s_n-\sigma_i |$, $i=1, 2$. Suppose that $\sigma_1<\sigma_2$ in the highest intensity interval. Let
\begin{equation}
\label{eq33}
\begin{aligned}
\mathcal{A}=\chi_{k}^{2}\left ( \frac{k\cdot s_n^2}{\sigma_2^2\left ( m_n \right )} \right ),\\
\mathcal{B}=\chi_{k}^{2}\left ( \frac{k\cdot s_n^2}{\sigma_1^2\left ( m_n \right )} \right ),
\end{aligned}
\end{equation}
where $k$ denotes the freedom of chi-square distribution. Note that the probabilities of $\mathcal{A}$ and $\mathcal{B}$ rely on the ratio between the noise sample and NLF, which can be regarded as the equivalence of $\varepsilon _i$. By substituting (\ref{eq33}) into (\ref{eq8}) and (\ref{eq22}), we have
\begin{equation}
\label{eq34}
\begin{aligned}
p\left ( s^{2}\in \sigma _{1}\left ( m \right )|s^{2} \right )
&=\frac{\mathcal{B}\cdot p\left ( \sigma _{1}\left ( m \right )  \right )}{\mathcal{A}\cdot p\left ( \sigma _{2}\left ( m \right )  \right )+\mathcal{B}\cdot p\left ( \sigma _{1}\left ( m \right )  \right )},\\
p\left ( s^{2}\in \sigma _{2}\left ( m \right )|s^{2} \right )
&=\frac{\mathcal{A}\cdot p\left ( \sigma _{2}\left ( m \right )  \right )}{\mathcal{A}\cdot p\left ( \sigma _{2}\left ( m \right )  \right )+\mathcal{B}\cdot p\left ( \sigma _{1}\left ( m \right )  \right )},
\end{aligned}
\end{equation}
where $p\left ( \sigma _{i}\left ( m \right )\right)$ is the $p\left ( s_n^2\in\sigma _{i}\left ( m \right )\right)$ for short.
First, we will study the situation of the noise samples on the one side of an NLF curve. With respect to the noise samples on the one side of the NLF $\sigma_1$, (that is, these noise samples lie under $\sigma_1$), the probabilities calculated by (\ref{eq34}) are only dependent on the numerical relation represented by $\mathcal{A}\cdot p(\sigma_2 (m)) $ and $\mathcal{B}\cdot p(\sigma_1 (m))$. Empirically, we have $p(\sigma_2 (m))<p(\sigma_1 (m))$. Obviously, $s_n^2/\sigma_2^2 (m_n )<s_n^2/\sigma_1^2 (m_n )$; and therefore, we have $\mathcal{A}<\mathcal{B}$ based on the monotonicity of probability density function (pdf) for chi-square distribution. Hence,
\begin{equation}
\label{eq35}
\mathcal{A}\cdot p\left ( \sigma_2\left ( m \right ) \right )<\mathcal{B}\cdot p\left ( \sigma_1\left ( m \right ) \right ).
\end{equation}
Referring to (\ref{eq35}), it is known that $p(s^2\in\sigma_1 (m)|s^2 )>p(s^2\in\sigma_2 (m)|s^2 )$ from (\ref{eq34}). This outcome means noise samples on the one side of the NLF $\sigma_1$ are regarded as pristine, which is also consonant to the reality. Concerning distance-based refinement, because $\varepsilon_1<\varepsilon_2$ or even $\varepsilon_1\ll \varepsilon_2$, noise samples on the one side of the NLF $\sigma_1$ are also considered as pristine, which is consistent with probability-based method. As for the noise samples on the one side of the NLF $\sigma_2$, (that is, these noise samples lying above $\sigma_2$), the same analysis yields a similar result, that they are classified as tampered.

\begin{figure}[!t] 
	\centering
	\includegraphics[width=1.5in]{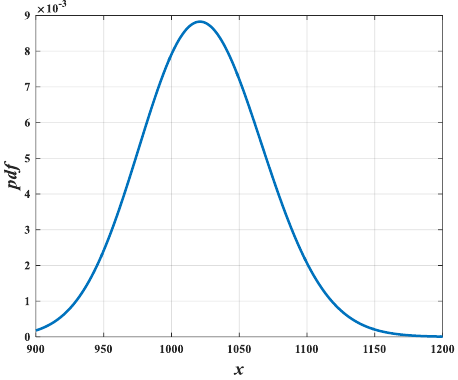}
	\caption{ Probability density function of chi-square distribution with degrees of freedom at the $k=32^2-1$.}
	\label{fig14}
	\vspace{-0.5cm}
\end{figure}

In the following, we would like to present an analysis of some detection results with respect to the noise samples that are closer to NLF obtained from the splicing region, between two curves. Suppose that detection based on the probability model is wrong; that is, $ p(s^2\in\sigma_1 (m)|s^2 )<p(s^2\in\sigma_2 (m)|s^2 )$, where $s^2$ is actually a noise variance calculated from an authentic region that is may be textural. Considering this scenario, we have
\begin{equation}
\label{eq36}
\frac{\mathcal{B}}{\mathcal{A}}\cdot \frac{p\left ( \sigma_1\left ( m \right ) \right )}{p\left ( \sigma_2\left ( m \right ) \right )}<1\Leftrightarrow 0<\frac{\mathcal{B}}{\mathcal{A}}<\frac{p\left ( \sigma_2\left ( m \right ) \right )}{p\left ( \sigma_1\left ( m \right ) \right )}.
\end{equation}
Empirically, according to multiple test images, we have
\begin{equation}
\label{eq37}
p\left ( \sigma_1\left ( m \right )\right )\approx \left ( 8\sim 15 \right )\cdot p\left ( \sigma_2\left ( m \right )\right ).
\end{equation}
For simplicity of analysis, we assume that $p\left ( \sigma_1\left ( m \right )\right )=10\times p\left ( \sigma_2\left ( m \right )\right )$. Then, by substituting it into (\ref{eq36}), we have
\begin{equation}
\label{eq38}
0<\frac{\mathcal{B}}{\mathcal{A}}<\frac{1}{10}.
\end{equation}
The result from (\ref{eq38}) demonstrates that if the probabilities calculated using (\ref{eq33}) satisfy (\ref{eq38}), some textural patches of authentic region are likely to be identified as tampered, which is incorrect. Next, we will describe this scenario in detail. For the sake of analysis, we take $k=32^2-1$ as the freedom of the chi-square distribution, and the illustration of the pdf is shown in Fig. \ref{fig14}.
Let $\sigma_2 (m_n )=s_n+\varepsilon_2$ and $\sigma_1 (m_n )=s_n-\varepsilon_1$. Two cases will be discussed for the noise samples lying between two NLF curves.\\
\emph{Case 1}: $\varepsilon_2$ is so tiny as to be even negligible compared with $\varepsilon_1$, i.e., the noise sample is very close to the NLF $\sigma_2 (m_n )$. Then
\begin{equation}
\label{eq39}
\begin{aligned}
\mathcal{A}&=\chi _k^2\left ( k\cdot \frac{\left ( \sigma_1\left ( m_n \right )+\varepsilon _1 \right )^2}{\left ( \sigma_1\left ( m_n \right )+\varepsilon _1+\varepsilon _2 \right )^2} \right )\approx \chi _k^2\left ( k \right ),\\
\mathcal{B}&=\chi _k^2\left ( k\cdot \left ( 1+\frac{\varepsilon _1}{\sigma_1\left ( m_n \right )} \right )^2 \right ).
\end{aligned}
\end{equation}
Since $k=32^2-1$, if (\ref{eq38}) is satisfied, we have an approximation according to Fig. \ref{fig14}, as follows:
\begin{equation}
\label{eq40}
\left ( 1+\frac{\varepsilon _1}{\sigma_1\left ( m_n \right )} \right )^2>\frac{1122}{1021},
\end{equation}
that is,
\begin{equation}
\label{eq41}
\varepsilon _1>0.0483\cdot \sigma_1\left ( m_n \right ).
\end{equation}
The result from (\ref{eq41}) demonstrates a condition in which incorrect detection happens for textural patches whose noise estimation lies near the inaccurate NLF $\sigma_2 (m_n )$. Moreover, it can also be seen from the formula that the detection results are likely to be unreliable for the sample points around the intersection of the two curves, because (\ref{eq41}) is incidental for these sample points. By the way, the intersection of two curves may result from an inaccurate recovery of NLF $\sigma_2 (m_n )$.\\
\emph{Case 2}:  $\varepsilon_2$ cannot be regarded as negligible compared with $\varepsilon_1$. Considering the formula of pdf for the chi-square distribution
\begin{equation}
\label{eq42}
f\left ( x;k \right )=\left\{\begin{matrix}
&\frac{x^{k/2-1}e^{-x/2}}{2^{k/2-1}\Gamma \left ( k/2 \right )}&,\,\, x>0,\\
&0&,\,\,{\rm otherwise}.
\end{matrix}\right.
\end{equation}
Assuming that $\sigma_2 (m_n )=2\cdot \sigma_1 (m_n )$ to simplify the numerical analysis. By substituting (\ref{eq42}) into (\ref{eq38}), then
\begin{equation}
\label{eq44}
\begin{aligned}
\frac{\mathcal{B}}{\mathcal{A}}&=\frac{\left ( \frac{k\cdot s_n^2}{\sigma_1^2\left ( m_n \right )} \right )^{k/2-1}e^{-\frac{1}{2}\left ( \frac{k\cdot s_n^2}{\sigma_1^2\left ( m_n \right )} \right )}}{\left ( \frac{k\cdot s_n^2}{\sigma_2^2\left ( m_n \right )} \right )^{k/2-1}e^{-\frac{1}{2}\left ( \frac{k\cdot s_n^2}{\sigma_2^2\left ( m_n \right )} \right )}}\\
&=\left ( \frac{\sigma_2\left ( m_n \right )}{\sigma_1\left ( m_n \right )} \right )^{k-2}e^{-\frac{k\cdot s_n^2}{2}\left ( \frac{1}{\sigma_1^2\left ( m_n \right )}-\frac{1}{\sigma_2^2\left ( m_n \right )} \right ) }\\
&<\frac{1}{10}.
\end{aligned}
\end{equation}
Since $k=32^2-1$ and $\sigma_2 (m_n )=2\cdot \sigma_1 (m_n )$, we can know from (\ref{eq44}) that
\begin{equation}
\label{eq45}
\varepsilon _1>0.36\cdot \sigma_1\left ( m_n \right ).
\end{equation}

The results of (\ref{eq41}) and (\ref{eq45}) both indicate that conditional probability-based method is likely to present an incorrect clue for the detection only if the distance between samples and an NLF satisfy a certain condition. In summary, it is necessary to modify the method accordingly, based on conditional probability model.

\ifCLASSOPTIONcompsoc
  \section*{Acknowledgments}\label{acknowledgements}
\else
  \section*{Acknowledgment}
\fi

The authors would like to thank Dr B. Liu for his kind help of code implementation and anonymous reviewers for their valuable suggestions which helped to improve this paper. \emph{Our code will be released after the acceptance of this paper.}

\ifCLASSOPTIONcaptionsoff
  \newpage
\fi



%

%
\begin{IEEEbiographynophoto}{Mian Zou}
	received B.S. degree in electrical engineering from Hefei University of Technology, China, in 2018. He is currently pursuing an M.S. degree in electrical engineering from University of Shanghai for Science and Technology, China. His research interests include image processing and multimedia security.
\end{IEEEbiographynophoto}

\begin{IEEEbiographynophoto}{Heng Yao}
	received the B. Sc. degree from Hefei University of Technology, China, in 2004, the M. Eng. degree from Shanghai Normal University, China, in 2008, and the Ph. D. degree in signal and information processing from Shanghai University, China, in 2012. Since 2012, he has been with the faculty of the School of Optical-Electrical and Computer Engineering, University of Shanghai for Science and Technology, where he is currently an Associate Professor. His research interests include multimedia security, image processing, and pattern recognition. He has contributed more than 40 international peer-reviewed journal papers.
\end{IEEEbiographynophoto}
\begin{IEEEbiographynophoto}{Chuan Qin}
	received the B.S. degree in electronic engineering and the M.S. degree in signal and information processing from the Hefei University of Technology, Anhui, China, in 2002 and 2005, respectively, and the Ph.D. degree in signal and information processing from Shanghai University, Shanghai, China, in 2008. Since 2008, he has been with the Faculty of the School of Optical-Electrical and Computer Engineering, University of Shanghai for Science and Technology, where he is currently a Professor. He was with Feng Chia University, Taiwan, as a Post-Doctoral Researcher, from 2010 to 2012. His research interests include image processing and multimedia security. He has published over 130 papers in these research areas.
\end{IEEEbiographynophoto}
\begin{IEEEbiographynophoto}{Xinpeng Zhang}
received the B.S. degree in computational mathematics from Jilin University, China,	in 1995, and the M.E. and Ph.D. degrees in communication and information system from Shanghai University, China, in 2001 and 2004, respectively. Since 2004, he has been with the Faculty of the School of Communication and Information Engineering, Shanghai University, where he is currently a Professor. He was with the State University of New York at Binghamton as a Visiting Scholar from 2010 to 2011 and with the Konstanz University as an Experienced Researcher small sponsored by the Alexander von Humboldt Foundation from 2011 to 2012. He is currently with the Faculty of the School of Computer Science, Fudan University. His research interests include multimedia security, image processing, and digital forensics. He has published over 200 papers in these areas. He has served as an Associate Editor for the IEEE Transactions on Information Forensics and Security from 2014 to 2017.
\end{IEEEbiographynophoto}




\end{document}